\RequirePackage{fix-cm}
\documentclass[smallextended]{svjour3}       
\smartqed  
\usepackage{graphicx}
\usepackage[ruled]{algorithm2e}
\usepackage{booktabs}
\usepackage{amssymb}
\usepackage{amsmath} 
\usepackage{latexsym}
\usepackage{hyperref}
\hypersetup{hypertex=true,
	colorlinks=true,
	linkcolor=blue,
	anchorcolor=blue,
	citecolor=blue
}
\usepackage{enumerate}
\usepackage{marvosym}

\begin{document}

\title{A novel meta-learning initialization method for physics-informed neural networks
}

\author{Xu Liu        \and
        Xiaoya Zhang \and
        Wei Peng \and
        Weien Zhou \and
        Wen Yao\textsuperscript{\Letter}
}

\institute{\Letter Wen Yao \at
              National Innovation Institute of Defense Technology, Chinese Academy of Military Science, 100071, Beijing. \\
             \email{yaowen@nudt.edu.cn}           
}

\date{Received: date / Accepted: date}

\maketitle
 
\begin{abstract}
Physics-informed neural networks (PINNs) have been widely used to solve various scientific computing problems. However, large training costs limit PINNs for some real-time applications. Although some works have been proposed to improve the training efficiency of PINNs, few consider the influence of initialization. To this end, we propose a New Reptile initialization based Physics-Informed Neural Network (NRPINN). The original Reptile algorithm is a meta-learning initialization method based on labeled data. PINNs can be trained with less labeled data or even without any labeled data by adding partial differential equations (PDEs) as a penalty term into the loss function. Inspired by this idea, we propose the new Reptile initialization to sample more tasks from the parameterized PDEs and adapt the penalty term of the loss.
The new Reptile initialization can acquire initialization parameters from related tasks by supervised, unsupervised, and semi-supervised learning. Then, PINNs with initialization parameters can efficiently solve PDEs.
Besides, the new Reptile initialization can also be used for the variants of PINNs.
Finally, we demonstrate and verify the NRPINN considering both forward problems, including solving Poisson, Burgers, and Schr\"odinger equations, as well as inverse problems, where unknown parameters in the PDEs are estimated. Experimental results show that the NRPINN training is much faster and achieves higher accuracy than PINNs with other initialization methods.

\end{abstract}

\section{Introduction}
\label{intro}
Recently, neural networks (NNs) have been successfully applied to the content of solving partial differential equations (PDEs) \cite{brink2020neural,avrutskiy2020neural,tartakovsky2018learning,he2020physics,chen2020physics}.
In particular, the physics-informed neural networks (PINNs) \cite{raissi2019physics,raissi2020hidden} is a general framework for solving both forward and inverse problems of PDEs. It is well suited for solving complex PDE-based physics problems in various domains, such as materialogy \cite{shukla2020physics}, medical diagnosis \cite{sahli2020physics} and hidden fluid mechanics \cite{mao2020physics}, elastodynamics \cite{raissi2019deep}, etc. 
PINNs encode the governing equation into loss function by adding a penalty term to constrain the space of admissible solutions. 
The problem of solving PDEs is transformed into an optimization problem of minimizing the loss function. 
Generally, training of a deep neural network model requires a vast number of labeled data, 
but the solutions of PDEs can be learned through a PINN model with less labeled data \cite{raissi2019physics,zhang2020physics,sun2020physics} or even without any labeled data \cite{zhu2019physics,sun2020surrogate}.
Besides, compared to numerical methods, such as finite difference method \cite{narasimhan1976integrated},  wavelets method \cite{kumar2020efficient,kumar2021fractional,kumar2021wavelet} and laplace transform method \cite{kumar2014new}, the PINN is a mesh-free approach by utilizing \textit{automatic differentiation} \cite{baydin2017automatic} and avoids the curse of dimensionality \cite{poggio2017and,grohs2018proof}. 

Large training cost is one of the main limitations of PINNs to run in real-time for solving real-life applications. Therefore, it is crucial to improve its training efficiency without sacrificing the performance.
To improve the training efficiency of PINNs, D.Jagtap et al. \cite{jagtap2020adaptive,jagtap2019locally} introduced a scalable parameter in the activation function and it can be optimized during the optimization process. 
Inspired by works of D.Jagtap, Peng et al. \cite{peng2020accelerating} proposed a variant called Prior Dictionary based PINNs to capture features provided by prior dictionaries. 
By modifying the structure of NNs, Sitzmann et al. \cite{sitzmann2020implicit} used periodic activation functions for implicit neural representations and Tancik et al. \cite{tancik2020fourier} added a simple Fourier feature mapping in the input of NNs to learn high-frequency functions in low-dimensional problem domains, which further speeds up the convergence. For accelerating the training of long-time integration, Meng et al. \cite{meng2020ppinn} proposed a parareal PINN to decompose a long-time problem into many independent short-time problems.
D.Jagtap et al. \cite{jagtap2020conservative,jagtap2020extended} proposed conservative PINNs and extended PINNs, where the computational domains are decomposed to achieve the speedup. 
A similar approach was proposed by Dwivedi et al. \cite{dwivedi2019distributed}, where the authors decomposed the computational domain to many regular subdomains and installed a PINN in each subdomain.
Those works mostly focus on modifying the structure of NNs or decomposing the computational domains. However, they have not considered the effect of network initialization for training efficiency and predicted solutions.

Chakraborty et al. \cite{chakraborty2020transfer} and Goswami et al. \cite{goswami2020transfer} introduced transfer learning to initialize PINNs for dealing with multi-fidelity problems and brittle fracture problem, respectively. 
But they have not considered designing efficient initialization algorithms for accelerating the training of PINNs. 
Recently, the meta-learning algorithms have received more and more attention on initialization \cite{rajeswaran2019meta,smith2009cross,finn2017meta,finn2018probabilistic}. 
In particular, the Model-Agnostic Meta-Learning (MAML) is an important meta-learning algorithm for the initialization. MAML was proposed by Finn et al. \cite{finn2017model} to obtain the initialization parameters of NNs through training a variety of learning tasks. 
Although there is only a small amount of labeled training data, the model with MAML initialization converges well.
Based on MAML, the Reptile algorithm proposed by Nichol et al. \cite{nichol2018first} only uses the first-order information and achieves similar performance to the MAML. 
Compared to the MAML, the Reptile algorithm uses less computation and memory resources. 
The Reptile algorithm depends on labeled data to acquire initialization parameters through supervised learning. 
The amount of labeled data is limited in real-life applications, so the acceleration effect of the Reptile initialization is restricted. 
But PINNs are used with less labeled data or even without any labeled data. Inspired by this idea, the Reptile algorithm is promising to be extended for unsupervised and semi-supervised learning, which further improves the acceleration effect of initialization.

Initialization for PINNs is ignored in most works, where most researchers employed Xavier initialization \cite{glorot2010understanding}. But a good initialization can endow PINNs with a good start to achieve fast convergence and improve accuracy. 
How to obtain a good initialization through modifying the Reptile algorithm is a problem to be discussed. 
In this paper, we explore those questions and our main contributions are:

\begin{enumerate}[1)]
	\item PINNs add PDEs as a penalty term into the loss function and then can be learned with less labeled data or even without any labeled data. Inspired by this idea, we propose the new Reptile initialization to sample more tasks from the parameterized partial differential equations and adapt the penalty term of the loss.
	Faced with only labeled data, no labeled data, or data both with and without labels, the new Reptile initialization acquires initialization parameters through supervised, unsupervised, and semi-supervised learning, further improving the acceleration effect of initialization.
	\item Based on the new Reptile initialization, we propose a new Reptile initialization based physics-informed neural network (NRPINN) algorithm to achieve fast convergence and high accuracy for PINNs. The advantages of using the NRPINN to approximate solutions of PDEs can be summarized as follows: (i) we use NRPINN to obtain the surrogate model of solutions. Forward evaluations of the surrogate model are extremely effective particularly for many-query and real-time applications; (ii) the NN is differentiable and derivative information can be easily obtained via automatic differentiation, which is a profitable feature for control or optimization problems. (iii) the training of the NRPINN is much faster and achieves higher accuracy than PINNs with other initialization methods.
	Besides, the new Reptile initialization can also be used for the variants of PINNs, such as PINNs with adaptive activation function \cite{jagtap2020adaptive,jagtap2019locally}.
	\item Based on the NRPINN algorithm, we study forward problems, e.g. Poisson, Burgers, and Schr\"odinger equations, as well as inverse problems, where unknown parameters in the PDEs are estimated. In addition to studying the effect of noisy data for prediction, we use the new Reptile initialization for a variant structure of PINNs.
\end{enumerate}

The paper is structured as following sections. Problem statement and related work are presented in section \ref{problem}.
In section \ref{sec2}, after briefly introducing the key concepts of PINNs and Reptile, we present the NRPINN algorithm in detail. 
Numerical analysis of forward and inverse problems is conducted in section \ref{sec3}, where we discuss the results of PINNs with different initialization methods and the NRPINN.
The conclusions of the paper are in Section \ref{sec4}.

\section{Problem statement and related works}
\label{problem}
\subsection{Problem statement}
Generally, the parametrized and nonlinear PDEs is given by
\begin{equation}
\label{problem_eq}
\begin{aligned}
u_{t}+\mathcal{N}[u ; \lambda] &=0, x \in \Omega, t \in[0, T] \\
\text { B.C. }: \mathcal{B}(u, x) &=0, x \in \partial \Omega, t \in[0, T] \\
\text { I.C. }: u(0, x) &=u_{0}(0, x), x \in \partial \Omega,
\end{aligned}
\end{equation}
where $\mathcal{N}[; \lambda]$ is a nonlinear operator with the parameter $\lambda$, $\Omega$ is the physical domain and $\mathcal{B}(\cdot)$ is the boundary condition. For example, the nonlinear operator of the one-dimensional Burgers equation is given as $\mathcal{N}[u ; \lambda]=u^{2} / 2-\lambda u_{x}$, where the parameter $\lambda$ is the viscosity coefficient.

In this work, we mainly consider solving the forward problems as well as inverse problems. The forward problem is solutions of PDEs to be inferred with the fixed parameter $\lambda$. For inverse problems, the unknown parameter $\lambda$ is learned from the observed data. In addition, the solutions can be also inferred.
                                                                 
\subsection{Related works}
Recently, PINNs have been proven to be promising for solving PDEs or PDE-based systems \cite{sirignano2018dgm,jagtap2020adaptive,sun2020surrogate}. However, training efficiency and prediction accuracy of PINNs are still a tremendous challenge. A few works focused on modifying the structure of NNs to speed up the convergence. D.Jagtap et al. \cite{jagtap2020adaptive,jagtap2019locally} employed a scalable hyper-parameter in the activation function, which changes dynamically during the process of optimizing the loss function. The PINN with the adaptive activation function had a more desirable feature for speeding up the convergence rate and improving the solution accuracy. KIM et al. \cite{kim2020fast} presented a fast and accurate PINN ROM with a nonlinear manifold solution representation. The structure of NNs included two part, namely encoder and decoder. They trained a shallow masked encoder, where training data was from the full order model simulations. Then, the decoder was used as the nonlinear manifold solution representation. Merging prior information in the structure of NNs, Peng et al. \cite{peng2020accelerating} proposed a Prior Dictionary based PINNs to capture features provided by dictionaries so as to speed up the convergence.

Some works were to decompose the computational domain for speeding up the convergence. D. Jagtap et al. \cite{jagtap2020conservative,jagtap2020extended} proposed conservative PINNs and extended PINNs to decompose the computational domain to several discrete sub-domains, where a shallow PINN was employed in each sub-domain. Inspired by works of D. Jagtap, Shukla et al. \cite{shukla2021parallel} developed a distributed training framework for PINNs, where domain decomposition methods were used in space and time-space. The distributed framework combined the advantage of conservative PINNs and extended PINNs so as to accelerate the convergence. For solving the PDEs with long time integration, the time-space domain might become very large so that the training cost of NNs would become extremely expensive. To this end, Meng et al. \cite{meng2020ppinn} proposed a parareal PINN to solve the long-time problem. They used a fast coarse-grained solver to decompose the long-time domain into many discrete short-time domains. Training several PINNs with many small-data sets was much faster than a PINN with a large-data set. The parareal PINN could achieve a significant speedup for PDEs with long-time integration. Combining domain decomposition and projection onto space of high-order polynomials, Kharazm et al. \cite{kharazmi2021hp} introduced hp-variational PINNs to divide the computational space into the trial space and test space. Trial space was the space of NNs and test space represented the space of high order polynomials. The hp-variational PINNs learned from trial space and test space to accelerate the convergence and improve the solution accuracy. 

Other works introduced transfer learning to accelerate the training of PINNs. Many physics systems could obtain high-fidelity and low-fidelity data. Generally, high-fidelity data was a large-data and low-fidelity data was a small-data set. To best use of two data sets, Chakraborty et al. \cite{chakraborty2020transfer} proposed a novel multi-fidelity PINN. The low-fidelity data was first used to train a low-fidelity PINN. Then, they utilized high-fidelity data to obtain high-fidelity PINN by fine-tuning low-fidelity PINN. The multi-fidelity PINN provided a novel method to extract useful information from both low-fidelity and high-fidelity data. To solve brittle fracture problems, Goswami et al. \cite{goswami2020transfer} introduced transfer learning to re-train the NN partially, instead of re-training the complete network. With this training method, the training efficiency was significantly enhanced.
However, these methods ignored the acceleration effect of initialization. In most works, these existing initialization methods, such as Xavier and random initialization, did not use prior information. Therefore, how to use the prior information to endow PINNs with a good initialization is an open issue.

\section{Methodology}
\label{sec2}
\subsection{Physics-informed neural networks}
\begin{figure}[!h]
	\centering
	\includegraphics[scale=0.50]{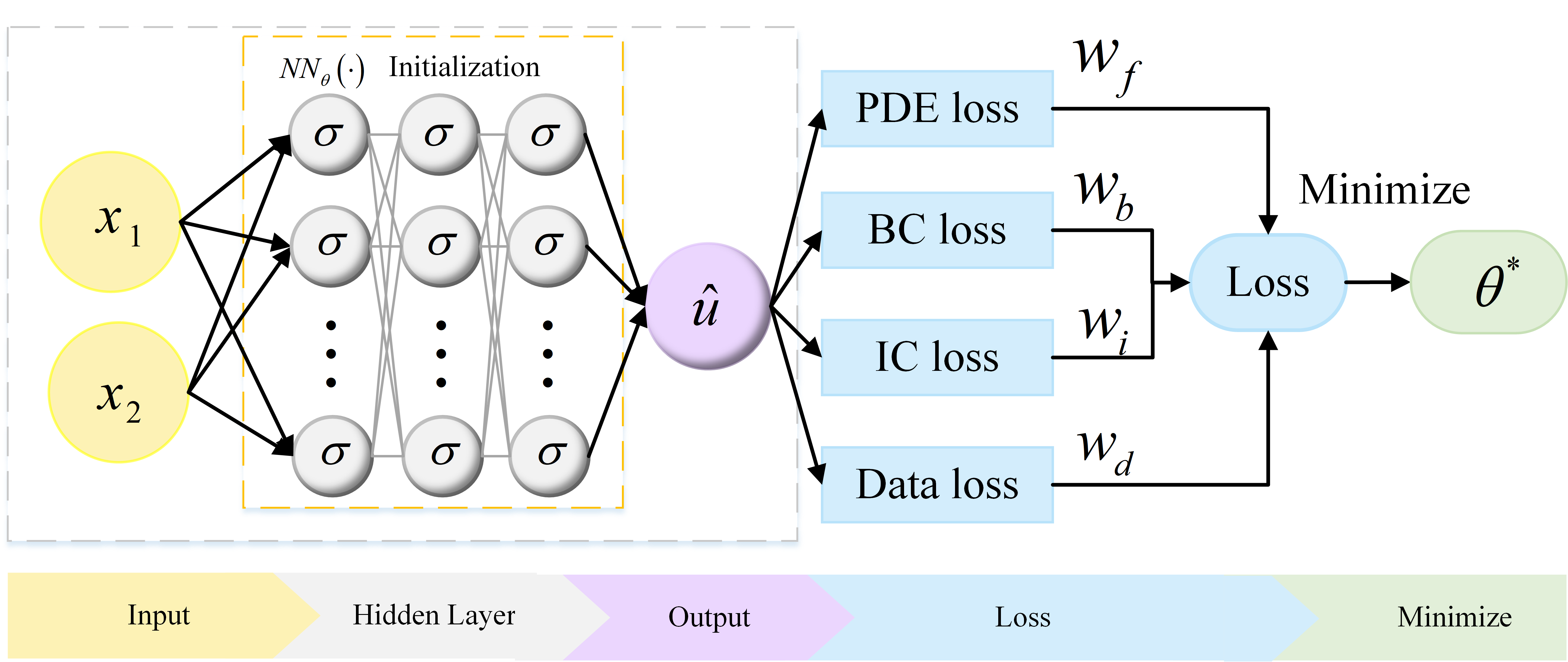}
	\caption{The schematic of a PINN.}
	\label{PINN}
\end{figure}

In this part, we first provide a brief overview of PINNs. Fig.\ref{PINN} is the schematic of a PINN. 
Let $N_{\theta}(\cdot)$ be a NN of depth $D$, where $\theta$ denotes the vector of initialization of the NN. The NN has an input layer, $D-2$ hidden layers and an output layer, which is a mapping from $\mathbb{R}^{d}$ into $\mathbb{R}^{N}$. 
We employ a Multi-Layer Perception (MLP) with the activation function $\sigma(\cdot)$.

To measure the difference between the NN and the physics constraints (governing equation and boundary condition), the left-hand-side of Eq.(\ref{problem_eq}) is defined as $f(t,x)$, which is given by
\begin{equation}
f(t,x):=u_{t}+\mathcal{N}[u].
\end{equation}

The loss function includes partial differentiable structure loss (PDE loss), boundary value condition loss (BC loss), initial value condition loss (IC loss), and true value condition (Data loss) which are defined as
\begin{equation}
\label{PDE_loss}
\mathcal{L}\left(\theta\right)=w_{f} \mathcal{L}_{PDE}\left(\theta; \mathcal{T}_{f}\right)+w_{i} \mathcal{L}_{IC}\left(\theta ; \mathcal{T}_{i}\right)+w_{b} \mathcal{L}_{BC}\left(\theta; \mathcal{T}_{b}\right)+w_{d} \mathcal{L}_{Data}\left(\theta; \mathcal{T}_{data}\right),
\end{equation}
where
\begin{equation}
\begin{aligned}
\mathcal{L}_{PDE}\left(\theta ; \mathcal{T}_{f}\right) &=\frac{1}{\left|\mathcal{T}_{f}\right|} \sum_{\mathbf{x} \in \mathcal{T}_{f}}\|f(t,\mathbf{x})\|_{2}^{2}, \\
\mathcal{L}_{IC}\left(\theta; \mathcal{T}_{i}\right) &=\frac{1}{\left|\mathcal{T}_{i}\right|} \sum_{\mathbf{x} \in \mathcal{T}_{i}}\|\hat{u}(\mathbf{x})-u(\mathbf{x})\|_{2}^{2}, \\
\mathcal{L}_{BC}\left(\theta; \mathcal{T}_{b}\right) &=\frac{1}{\left|\mathcal{T}_{b}\right|} \sum_{\mathbf{x} \in \mathcal{T}_{b}}\|\mathcal{B}(\hat{u}, \mathbf{x})\|_{2}^{2},\\
\mathcal{L}_{Data}\left(\theta; \mathcal{T}_{data}\right) &=\frac{1}{\left|\mathcal{T}_{data}\right|} \sum_{\mathbf{x} \in \mathcal{T}_{data}}\|\hat{u}(\mathbf{x})-u(\mathbf{x})\|_{2}^{2}.
\end{aligned}
\end{equation}
and $w_{f}$, $w_{i}$, $w_{b}$ as well as $w_{d}$ are predefined hyper-parameters of weights. In this manuscript, we choose equal weights. $\mathcal{T}_{f}$, $\mathcal{T}_{i}$, $\mathcal{T}_{b}$, and $\mathcal{T}_{data}$ denote the sets of residual points from PDE, initial value, boundary value, and real value, respectively. The $\hat{u}$ denotes the output of NN and $|\cdot|$ denotes the size of $\mathcal{T}$.

In the last step, the training procedure is to search for the best parameter $\theta^{*}$ of NN by minimizing the loss function $\mathcal{L}\left(\theta\right)$, where \textit{automatic differentiation} is used to minimize the highly nonlinear and non-convex loss function $\mathcal{L}\left(\theta\right)$.

\subsection{Reptile algorithm}
MAML and Reptile are important algorithms for meta-learning initialization. 
Compared to MAML, the Reptile algorithm \cite{nichol2018first} only performs stochastic gradient descent (SGD) \cite{bottou1991stochastic} or Adam \cite{kingma2014adam} (a gradient descent optimization algorithm) on each task once without twice like MAML, making less computation and memory resources, but the Reptile algorithm can achieve similar performance to MAML. 
The key of the Reptile algorithm is to obtain the initialization parameters so that the model can get the maximum performance of accuracy in the new task after updating the parameters through one or more gradient steps. 

A detailed description of the Reptile algorithm is shown in \textbf{Algorithm \ref{alg1}}.
The optimization problem of the Reptile is to find a set of initialization parameters $\phi$. 
The learner constructs the corresponding loss $L_{\tau}$ through a sampled task $\tau$. Each sampled task contains labeled data for classification or regression. 
For instance, a task might be a classification task from a labeled dataset with cats and dogs. 
Next, the learner finds parameters $\tilde{\phi}$ to minimize the loss $L_{\tau}$ after $k$ gradient steps. 
Then, the parameters are updated in the direction $\phi+\epsilon(\widetilde{\phi}-\phi)$.
In the last step, the parameters for Reptile initialization can be obtained by gradient updates under different tasks. The NN with initialization parameters for a new task can achieve the maximum performance of accuracy after a few gradient updates. The Reptile algorithm requires a large number of labeled data for supervised learning.
However, when the amount of labeled data is limited, the initialization effect for accelerating convergence is restricted.
\begin{algorithm*}[!h]
	\caption{\textbf{Reptile}}
	\label{alg1}
	{   Construct a NN $\hat{u}(x;\phi)$ with parameters $\phi$\\
		\For{Iteration=$1,2, \dots$}
		{
			Sample task $\tau$, corresponding to loss $L_{\tau}$ on weight vector $\tilde{\phi}$ \\
			Compute $\widetilde{\phi}=U_{\tau}^{k}(\phi)$, denoting k steps SGD or Adam\\
			Update $\phi \leftarrow \phi+\epsilon(\widetilde{\phi}-\phi)$
		}
		
	}
\end{algorithm*}

\subsection{The New Reptile initialization based Physics-Informed Neural Network(NRPINN)}
PINNs add PDEs as a penalty term into the loss function and then can be learned with less labeled data or even without any labeled data. Inspired by this idea, we propose the new Reptile initialization through modifying the task sampling process and the penalty term of the loss. The new Reptile initialization is extended from supervised learning to unsupervised and semi-supervised learning. To this end, we propose a new Reptile initialization based physics-informed neural network (NRPINN), which can achieve fast adapting and improve accuracy for the PINN. Our implementation is all done in Torch (version 1.4.0) using the Python application programming interface. According to the learning tasks in the new Reptile initialization, the NRPINN is divided into NRPINN with supervised learning (NRPINN-s), NRPINN with unsupervised learning (NRPINN-un) and NRPINN with semi-supervised learning (NRPINN-semi).
The detailed description of the NRPINN is shown in \textbf{Algorithm \ref{alg2}}. We consider a MLP represented by $\hat{u}(x ; \theta)$ with parameters $\theta$. 
The PDE to be solved is in the general form 
\begin{equation}
\label{PDE}
u_{t}+\mathcal{N}\left[u ; \lambda\right]=0, x \in \Omega, t \in[0, T],
\end{equation}
where $\mathcal{N}[\cdot ; \lambda]$ is a nonlinear operator parametrized by the known parameter $\lambda$, and $\Omega$ is a subset of $\mathbb{R}^D$. We summarize the prior information as two different information, namely high-order information and zero-order information. The two types of information can be represented as follows.

\begin{enumerate}[1)]
	\item The high-order information contains high-order derivative information of the solution $u$, which is described by
	\begin{equation}
	\label{high-order}
	u_{t}+\mathcal{N}[u ; \lambda]=0, \lambda \in P_{h}(\tau), x \in \Omega, t \in[0, T],
	\end{equation}
	where $\lambda$ is unknown and $P_{h}(\tau)$ denotes the set of $\lambda$ value. 
	In other words, we consider the known parameter $\lambda$ in Eq.(\ref{PDE}) as variables to obtain the Eq.(\ref{high-order}), which is called the high-order information of solution to be solved. Under high-order information, the governing equation with different $\lambda$ is considered as a task. 

	\item The zero-order information represents the solution $u_{\lambda}$ under different $\lambda$ can be obtained, where the solution $u_{\lambda}$ is composed of labeled data. The zero-order information is given by

	\begin{equation}
	\label{zero-order}
	u_{\lambda} \in P_{z}(\tau), x \in \Omega, t \in[0, T],
	\end{equation}
	
	where $P_{z}(\tau)$ represents the set of the labeled data under different parameter $\lambda$. For the zero-order information, the solution $u_{\lambda}$ under different parameter $\lambda$ is considered as a task.

\end{enumerate} 
The two types of information can be regarded as tasks of the new Reptile initialization. There are three ways to acquire parameters: (i) when only the zero-order information can be obtained, the new Reptile initialization acquires parameters through the tasks with labeled data. 
(ii) when only the high-order information can be obtained, the new Reptile initialization acquires parameters through the tasks with unlabeled data. 
(iii) when the two types of information is obtained, the new Reptile initialization acquires parameters through the tasks with labeled and unlabeled data.
\begin{algorithm*}[!h]
	\caption{\textbf{A new Reptile initialization based PINN}}
	\label{alg2}
	\LinesNumbered
	\KwIn{\\Initialize $\theta$, the vector of initial parameters of NN
		\\$P_{z}(\tau)$, the tasks set of zero-order information
		\\$P_{h}(\tau)$, the tasks set of high-order information\
		\\$N$ and $L$ denote the number of all tasks and updating all tasks
		\\$L_{z}$ and $L_{h}=L-L_{z}$ denote the number of tasks for supervised learning and unsupervised learning
	}
	\KwOut{\\  $\hat{u}(x;\theta)$ as the surrogate of the solution $u(x)$ }
	{   Construct a neural network $\hat{u}(x;\theta)$ with parameters $\theta$\\
		\For{$i \leftarrow 1:N$}{
			\For{$j \leftarrow 1:L_{z}+L_{h}$}{
				\eIf{$j < L_{z}$}
				{
					\textbf{Sample task $\tau_{j}$ from $P_{z}(\tau)$} \\
					Sample training points from task $\tau_{j}$\\
					Specify the loss $\mathcal{L}_{z}(\theta)$ between true value and predict value of NN \\
				}
				{
					\textbf{Sample task $\tau_{j}$ from $P_{h}(\tau)$} \\
					Specify training set $\mathcal{T}_{f}$ and $\mathcal{T}_{b}$ for PDE and boundary conditions from task $\tau_{j}$\\
					Specify the loss function $\mathcal{L}_{h}(\theta)$about PDE and boundary condition residuals by\\ summing the weighted $L^{2}$ norm\\
				}
				Compute $\widetilde{\theta}=U_{\tau}^{k}(\theta)$, denoting updating $\theta$ $k$ times by minimizing the loss function $\mathcal{L}_{z}(\theta)$ or $\mathcal{L}_{h}(\theta)$ through SGD or Adam 
			}
			Compute $\epsilon = \epsilon_{0}(1-j/N)$\\
			Update $\theta \leftarrow\theta+\epsilon(\widetilde{\theta}-\theta)$ to find the best parameters $\theta^{*}$
		}
	}
	Initialize the NN with parameters $\theta^{*}$\\
	Specify four training sets $\mathcal{T}_{f}$, $\mathcal{T}_{i}$, $\mathcal{T}_{b}$, and $\mathcal{T}_{f}$ for PDE, initial, boundary, and data conditions\\
	Specify the loss function $\mathcal{L}_{PINN}(\theta)$ about PDE, initial, and boundary condition constraints
	by summing the weighted $L^{2}$ norm\\
	Train the NN by minimizing the loss function $\mathcal{L}_{PINN}(\theta)$ until finding the best parameters $\theta^{**}$ \\
	Obtain $\hat{u}(x;\theta)$ by loading $\theta^{**}$
\end{algorithm*}

The procedure of the NRPINN algorithm is mainly divided into two parts, the new Reptile initialization and solving PDEs by PINNs. 
In the process of the new Reptile initialization, $L_{z}$ and $L_{h}$ represent the number of tasks for supervised and unsupervised learning, respectively. $L$ and $N$ represent the number of all tasks and updating all tasks. 
The new Reptile initialization is used for supervised, unsupervised, and semi-supervised learning. 
According to zero-order or high-order information, different types of tasks are obtained.
Under each task, the difference between the three learning types is whether labeled data is used in the task sampling process. The three ways to learn from the tasks are as follows.
\begin{enumerate}[1)]
	\vspace{0.3cm}
	\item For supervised learning when $L_{h}=0$, we sample task $\tau_{j}$ from $P_{z}(\tau)$ to obtain labeled data. Based on exact and predicted values, the loss function $\mathcal{L}_{z}(\theta)$ is confirmed. 
	\item For unsupervised learning when $L_{z}=0$, we sample task $\tau_{j}$ from $P_{h}(\tau)$ and specify training points set $\mathcal{T}_{f}$ and $\mathcal{T}_{b}$ from PDE and boundary conditions. Based on the training data, the loss function $\mathcal{L}_{h}(\theta)$ about PDE and boundary condition is specified.
	\item For semi-supervised learning, task $\tau_{j}$ can be obtained from $P_{z}(\tau)$ and $P_{h}(\tau)$, respectively. We consider the number of $L_{z}$ tasks for supervised learning and the remaining number $L_{h}=L-L_{z}$ of tasks is used for unsupervised learning.
\end{enumerate}

Then, for the same task, the parameter $\theta$ is updated $k$ times by minimizing the loss function through SGD or Adam, which is described by $\widetilde{\theta}=U_{\tau}^{k}(\theta)$. 
For different tasks, the parameter $\theta$ is updated in the direction $\widetilde{\theta}-\theta$ to find the best parameters $\theta^{*}$, and the update step size $\epsilon$ is a linear policy, $\epsilon=\epsilon_{0}(1-j / N)$.

The second part is the solving PDEs by PINNs. 
Through the new Reptile initialization, initialization parameters can be obtained. 
After loading the parameter $\theta^{*}$, the NN is restricted to satisfy the physics constraints. 
Then the training sets $\mathcal{T}_{f}$, $\mathcal{T}_{i}$, and $\mathcal{T}_{b}$ are specified for PDE, initial, boundary, and real data conditions. 
The loss $\mathcal{L}_{PINN}(\theta)$ by Eq.(\ref{PDE_loss}) is constructed to measure the discrepancy between the NN and the physics constraints.
In the last step, the training procedure is to search for the best parameter $\theta^{**}$ by minimizing the loss function $\mathcal{L}_{PINN}\left(\theta\right)$, where \textit{automatic differentiation} is used to minimize the highly nonlinear and non-convex loss function $\mathcal{L}_{PINN}\left(\theta\right)$.
It is worth noting that the new Reptile initialization is used for initialization part, instead of changing the structure of the NN. Therefore, the new Reptile initialization can also be flexibly used for the variants of PINNs, such as PINNs with adaptive activation \cite{jagtap2020adaptive}, parareal PINNs \cite{meng2020ppinn} and conservative PINNs \cite{jagtap2020conservative}, etc.

\section{Numerical experiments}
\label{sec3}

PDEs are the principle of describing many physical phenomena \cite{goufo2020similarities,kumar2020analysis,kumar2020study,kumar2020chaotic}. The NRPINN is a meshless method for PDE-based physics systems. To demonstrate and illustrate the performance of the NRPINN on different PDEs, we consider Poisson, Burgers, and Schr\"odinger equations, which have many real-life applications. Poisson equation is an elliptic PDE of broad utility in electrostatics, mechanical engineering, and theoretical physics \cite{fogolari2002poisson,bates1987glossary,pardoux2001poisson}. Burgers equation is a fundamental PDE that appears in various areas of applied mathematics, such as nonlinear acoustics \cite{hamilton1998nonlinear}, traffic flow \cite{may1990traffic}, fluid mechanics \cite{munson2013fluid}, etc. Schr\"odinger equation \cite{berezin2012schrodinger,veeresha2020fractional} is one of the fundamental equations of quantum mechanics \cite{griffiths2018introduction}, which combines the concept of matter wave and wave equation to describe the motion of microscopic particles. Using NRPINN to solve these PDEs is important for modeling the large and complex PDE-based physics problems.

In this section, we will first consider solving forward and inverse problems by the NRPINN over PINNs with different initialization methods, where Xavier initialization \cite{glorot2010understanding} is used by Raissi et al. \cite{raissi2019physics} and other initialization methods are widespread use. 
The forward problems include one-dimensional Poisson, two-dimensional Poisson, Burgers, and Schr\"odinger equations. 
The inverse problem includes Burgers equations, where the clean and noisy data is used to identify the unknown parameters in the governing equations. 
Besides, the new Reptile initialization is used for PINNs with adaptive activation function \cite{jagtap2020adaptive}.
\subsection{Poisson equation}
Poisson equation is represented by
\begin{equation}
\Delta \varphi=f,
\end{equation}
where $\Delta$ denotes Laplace operator and $f$ is penalty term.
\subsubsection{One-dimensional Poisson equation}
\label{One-dimensional Poisson equation}
For one dimensional problem, the zero-order information is given by
\begin{equation}
u_{\lambda}:=\zeta_{1} \sin \left(\eta_{1} x\right)+\zeta_{2} \cos \left(\eta_{2} x\right)-\zeta_{3} x+\eta_{3},
\end{equation}
where $\zeta_{i},\eta_{i} \sim \text { uniform }[0,2], \forall i=1,2,3$. 
$u_{\lambda}$ denotes the solution under different parameters.
Different tasks can be obtained from the zero-order information, and each task represents the solution under different parameters $\zeta_{i},\eta_{i}$.
The high-order information is given by
\begin{equation}
\frac{d^{2} u}{d x^{2}}=-\alpha^{2} \cdot \sin (\alpha x)-\beta^{2} \cdot \cos (\beta x),
\end{equation}
where $\alpha \sim \text { uniform }[0,1]$ and $\beta \sim$ uniform [0,2]. 
Different tasks can also be obtained from the high-order information, and each task represents the governing equation under different parameters.
In this case, we consider the following Poisson equation to be solved.
\begin{equation}
\begin{aligned}
\frac{d^{2} u}{d x^{2}} &=-0.49 \cdot \sin (0.7 x)-2.25 \cdot \cos (1.5 x), \\
u(-10) &=-\sin (7)+\cos (15)+1, \\
u(10) &=\sin (7)+\cos (15)-1,
\end{aligned}
\end{equation}
where $x \in(-10,10)$. The exact solution is smooth and has a linear term combining with two different frequency components, which is given by
\begin{equation}
u:=\sin (0.7 x)+\cos (1.5 x)-0.1 x.
\end{equation}

We use four hidden layers with 50 neurons in each layer. 
The NRPINN mainly has two steps. In the part of the new Reptile initialization, we consider three types of the new Reptile initialization: (i) for supervised learning, we randomly select 4,000 tasks from the zero-order information. 
Then, we sample 2,000 training points under each task, confirm loss function $\mathcal{L}_{z}(\theta)$, and use Adam to update the parameter $\theta$ 20 times with the learning rate of 0.001. 
(ii) for unsupervised learning, we randomly select 60 tasks from high-order information. 
Under each task, we sample 10,000 residual points from PDE as training points. 
Based on training points, the loss function $\mathcal{L}_{h}(\theta)$ is confirmed and Adam is used to update the parameter $\theta$ 1,000 times with the learning rate of 0.001. 
(iii) for semi-supervised learning, we select 100 tasks, of which half is used for supervised learning and the other half for unsupervised learning. 
In the part of solving PDEs by PINNs, the PINN with initialization parameters is used to solve the PDEs. 
We use 500 residual training points, two training data on the boundary, and 50 real data. 

\begin{figure*}[!h]
	\includegraphics[scale=0.31]{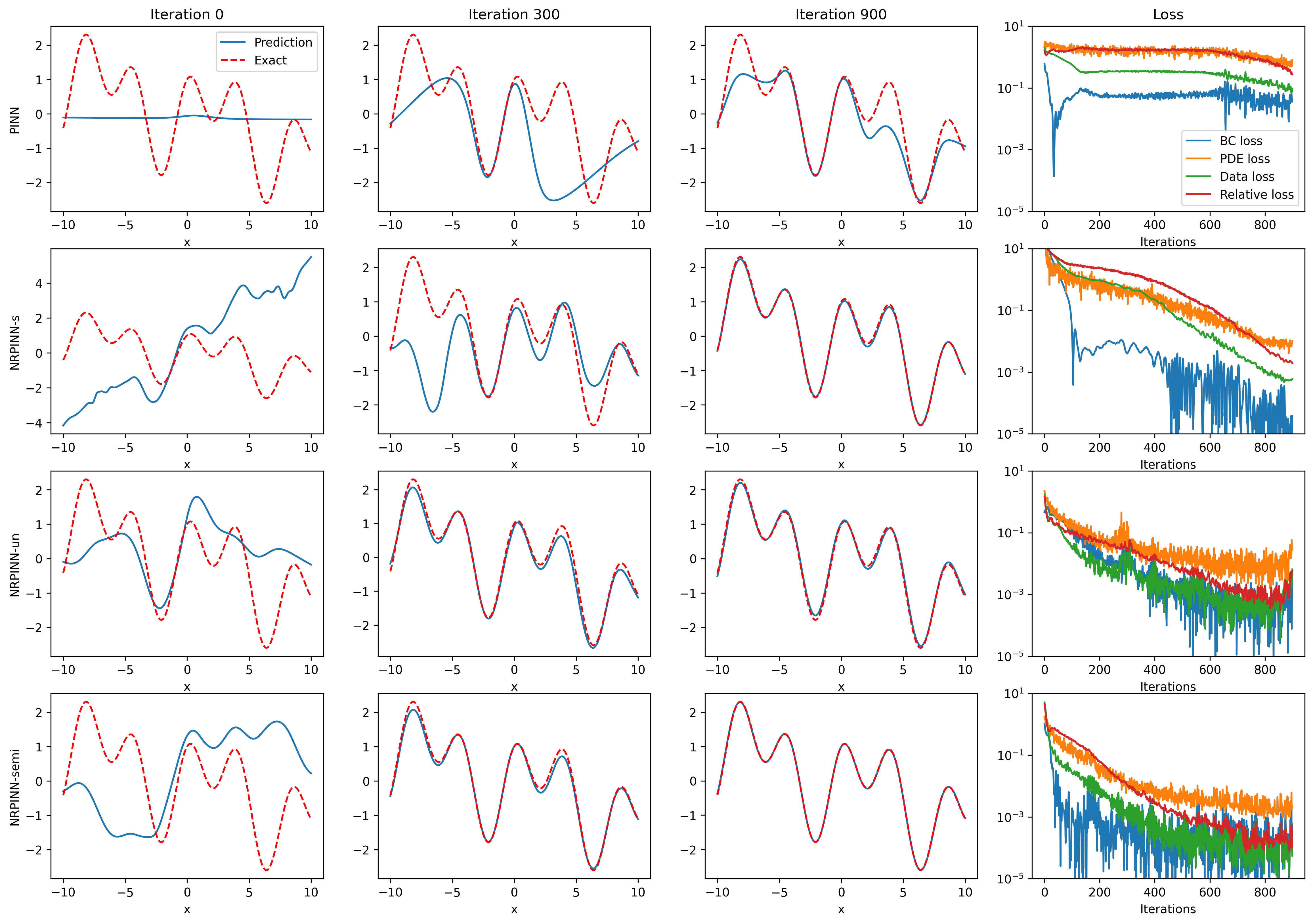}
	\caption{One-dimensional Poisson equation: The top row represents the result of the PINN with Xavier initialization and the last three rows represent the NRPINN-s, NRPINN-un, and NRPINN-semi, respectively. The first three columns represent the result after iteration 0, 300, and 900. The last column represents the loss versus the number of iterations.}
	\label{three_init}
\end{figure*}
\begin{figure*}[!t]
	\includegraphics[scale=0.34]{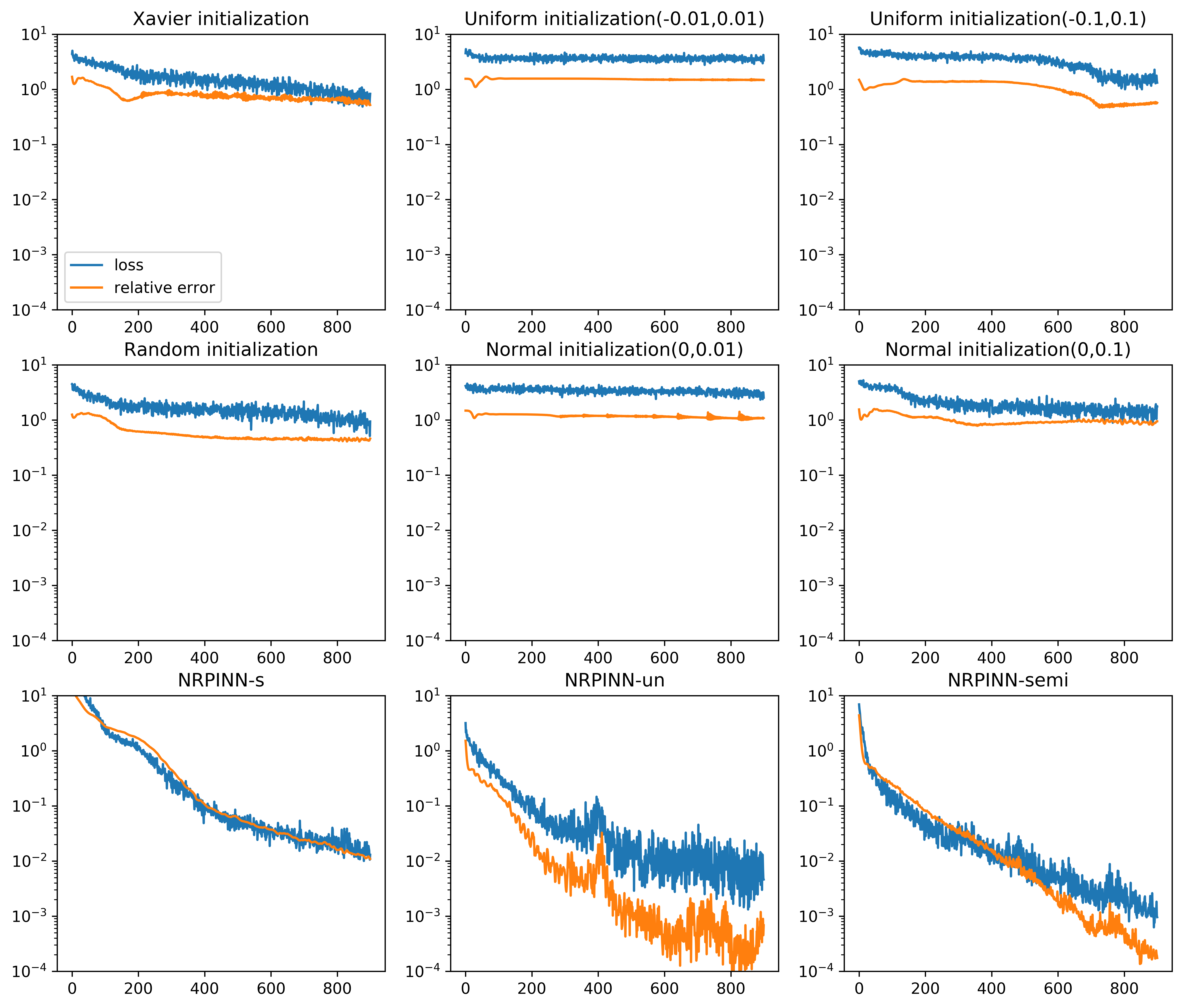}
	\caption{One-dimensional Poisson equation: The historical loss and relative error versus the number of iterations under different initialization methods. The first row presents Xavier initialization and two uniform distribution initialization with $U(-0.01,0.01)$ and $U(-0.01,0.1)$. The second row presents the PINN with random and normal distribution initialization ($\mathcal{N}\left(0,0.01\right)$ and $\mathcal{N}\left(0,0.1\right)$). The last row shows the historical loss of NRPINN-s, NRPINN-un, and NRPINN-semi, respectively.} 
	\label{Poisson_1d_ini}
\end{figure*}

Fig.\ref{three_init} (top row) shows the result of PINNs by Xavier initialization, where the loss falls very slowly and reaches  $1e-01$ till 900 iterations and predicted loss is still high. Fig.\ref{three_init} (the last three rows) show the predicted solution at various iterations as well as the loss versus the number of iterations by the NRPINN with supervised (NRPINN-s), the NRPINN with unsupervised (NRPINN-un), and the NRPINN with semi-supervised learning (NRPINN-semi), respectively. 
The NRPINN-s, NRPINN-un, and NRPINN-semi accurately capture wave features in the solution quickly and different types of losses are very slow within $300$ iterations. 
The predicted loss (red line) of the NRPINN-s, NRPINN-un, and NRPINN-semi has converged to $1e-05$, 
which is significantly better than $1e-01$ by the PINN with Xavier initialization.
Fig.\ref{Poisson_1d_ini} shows the history and relative loss with different initialization, where NRPINN-s, NRPINN-un, and NRPINN-semi outperform the PINN with other initialization methods. 
By contrast, the convergence performance of the PINN with uniform and normal distribution initialization is worst and the loss by Xavier initialization does not reach $1e-01$.

Table \ref{poisson_1d_rel} shows the MAEs between predicted solutions and exact solutions for different initialization methods after 900 iterations. From the table, it can be noted that the MAEs of NRPINN-s, NRPINN-un, and NRPINN-semi are far less than other initialization methods. In particular, the MAE of the NRPINN-un tends to be $5.1432e-04$, which is significantly better than $3.6731e-01$ by the PINN with the Xavier initialization.
Overall, the NRPINN can endow the PINN with a better initialization.

\begin{table}[!b]
	\caption{\label{poisson_1d_rel} One-dimensional Poisson equation: The MAEs of prediction solution with different initialization methods after 900 iterations. The best result of NRPINN is highlighted.}
	\begin{tabular}{l|l}
		\hline
		\multicolumn{1}{c|}{Initialization method} & \multicolumn{1}{c}{MAE} \\ \hline
		Xavier                                     & 3.6731e-01              \\ \hline
		Uniform(-0.01,0.01)                        & 1.7459                  \\ \hline
		Uniform(-0.1,0.1)                          & 1.7463                  \\ \hline
		Random                                     & 5.0123e-01              \\ \hline
		Normal(0,0.01)                             & 1.7400                  \\ \hline
		Normal(0,0.1)                              & 1.1109                  \\ \hline
		NRPINN-s                                    & 2.8012e-02              \\ \hline
		\textbf{NRPINN-un}                          & \textbf{5.1432e-04}     \\ \hline
		NRPINN-semi                                 & 6.2110e-04              \\ \hline
	\end{tabular}
\end{table}

\subsubsection{Two-dimensional Poisson equation}
For two dimensional Poisson equation, it is represented by 

\begin{equation}
\label{source_n}
\begin{aligned}
-\Delta u(x, y) &=\sum_{i=1}^{n} f\left(x, y ; a_{i}, b_{i}, c_{i}\right), \\
x, y \in \Omega &=[0,1] \times[0,1], \\
\left.u\right|_{\partial \Omega} &=0,
\end{aligned}
\end{equation}
where $n$ represents the number of heat source and the heat source is generated by the following scheme as follows.
\begin{equation}
\label{schem}
\begin{aligned}
f\left(x, y ; a_{i}, b_{i}, c_{i}\right)=& c_{i} \cdot \exp \left(-\frac{\left(x-a_{i}\right)^{2}+\left(y-b_{i}\right)^{2}}{0.01}\right),
\end{aligned}
\end{equation}
where $a_{i}$, $b_{i}\sim \text { uniform }[0.1,0.9]$ and $c_{i} \sim \text { uniform }[0.8,1.2]$. 

\begin{figure}[!h]
	\includegraphics[scale=0.18]{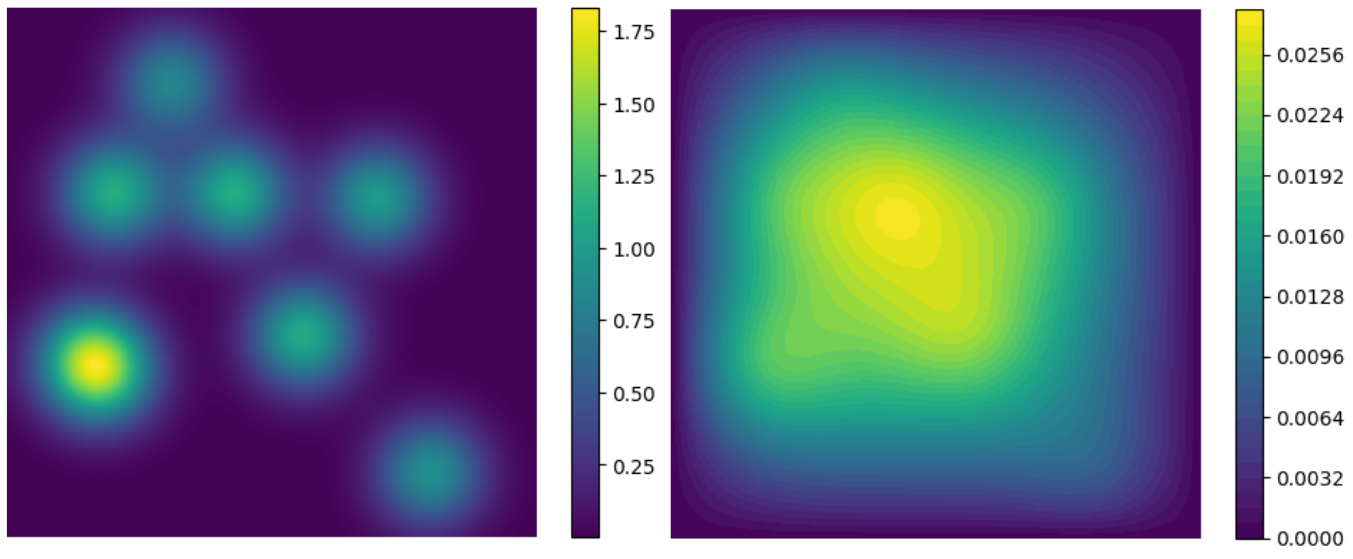}
	\caption{The eight heat sources locations (left) and the numerical solution (right), which is obtained by FEniCS }
	\label{8h}
\end{figure}

In this case, the PDE to be solved is when $n=8$ and its parameters are shown in Table \ref{location_8}. 
Its numerical solution can be obtained through FEniCS \cite{dupont2003fenics}, which is a popular open-source computing platform for solving PDEs. The numerical solution is shown in the Fig.\ref{8h}. 
The high-order information is the PDE with 1, 5 and 10 heat sources, which is given by
\begin{equation}
\label{1510}
\begin{aligned}
-\Delta u(x, y) &=\sum_{m=1,5,10} \sum_{i=1}^{m} c_{i} \cdot \exp \left(-\frac{(x-a_{i})^{2}+(y-b_{i})^{2}}{0.01}\right),
\end{aligned}
\end{equation}
where $a_{i}, b_{i} \sim \text { uniform }[0.1,0.9]$ and $c_{i} \sim \text { uniform }[0.8,1.2]$. 
Different tasks can be obtained from the high-order information, and each task represents the governing equation under different parameters.
The zero-order information is the set of the numerical solution under different parameters $a_{i}$, $b_{i}$ and $c_{i}$. 
Different tasks can be obtained from the zero-order information, and each task represents the solution of Eq.(\ref{1510}) under different parameters. The numerical solution is obtained through FEniCS.

\begin{table*}[!h]
	\caption{\label{location_8}The parameter of eight heat sources.}
	\begin{tabular}{c|c|c|c|c|c|c|c}
		\hline
		$a_{1}$ & $a_{2}$ & $a_{3}$  &  $a_{4}$& $a_{5}$&$a_{6}$ & $a_{7}$ &  $a_{8}$  \\
		\hline 
		0.15  &0.18& 0.20  & 0.31 & 0.43 & 0.56 &0.70 & 0.80   \\
		\hline
		$b_{1}$  & $b_{2}$ & $b_{3}$ &$b_{4}$& $b_{5}$&	$b_{6}$ & $b_{7}$  & $b_{8}$ \\
		\hline
		0.34 & 0.31 & 0.65 & 0.86 & 0.65 &	0.38 & 0.64  & 0.12  \\
		\hline
		$c_{1}$ & $c_{2}$ & $c_{3}$ & $c_{4}$& $c_{5}$&	$c_{6}$ & $c_{7}$  & $c_{8}$ \\
		\hline
		0.84 & 1.07 & 1.12 & 0.83& 1.12 &1.11 & 0.99  &0.91  \\
		\hline
	\end{tabular}
\end{table*}

We use four hidden layers with 100 neurons in each layer. 
In the new Reptile initialization part, we consider three types of new Reptile initialization: (i) for supervised learning, we assume that 100 tasks can be obtained from the zero-order information. 
Then, we sample 4,000 training points under each task, confirm loss function $\mathcal{L}_{z}(\theta)$, and use Adam to update the parameters of the NN 10,000 times with the learning rate of 0.001. 
(ii) for unsupervised learning, we randomly select 100 tasks from high-order information. Under each task, we sample 4,000 residual points from PDE as training points. 
Based on training points, the loss function $\mathcal{L}_{h}(\theta)$ is confirmed and Adam is used to update the parameter $\theta$ 10,000 times with the learning rate of 0.001. 
(iii) for semi-supervised learning, we select 100 tasks, of which half is used for supervised learning and the other half for unsupervised learning. 
In the part of solving PDEs by PINNs, the PINN with initialization parameters is used to solve the PDEs. 
We use 4,000 residual training points, 1,000 training data points on the boundary, and 0 real data. 

\begin{figure}[!h]
	\includegraphics[scale=0.28]{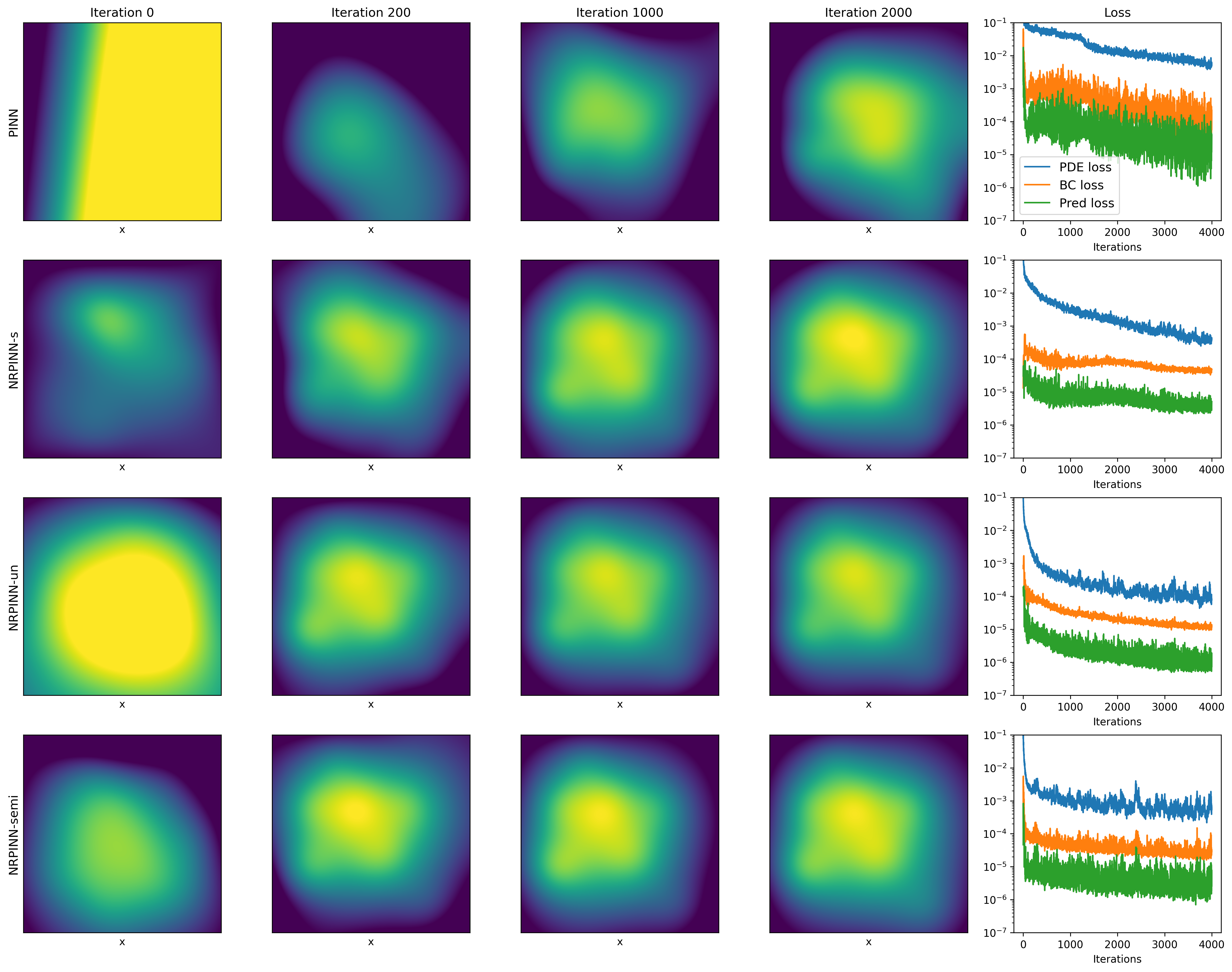}
	\caption{Two dimensional Poisson equation: The first row represents the result of PINNs with Xavier initialization and the last three row represent the NRPINN-s, NRPINN-un and NRPINN-semi, respectively. The first four columns represent the result after iteration 0, 200, 1000 and 2000. The last column represents the PDE, BC and Predicted loss history.}
	\label{8}
\end{figure}

\begin{figure}[!h]
	\includegraphics[scale=0.34]{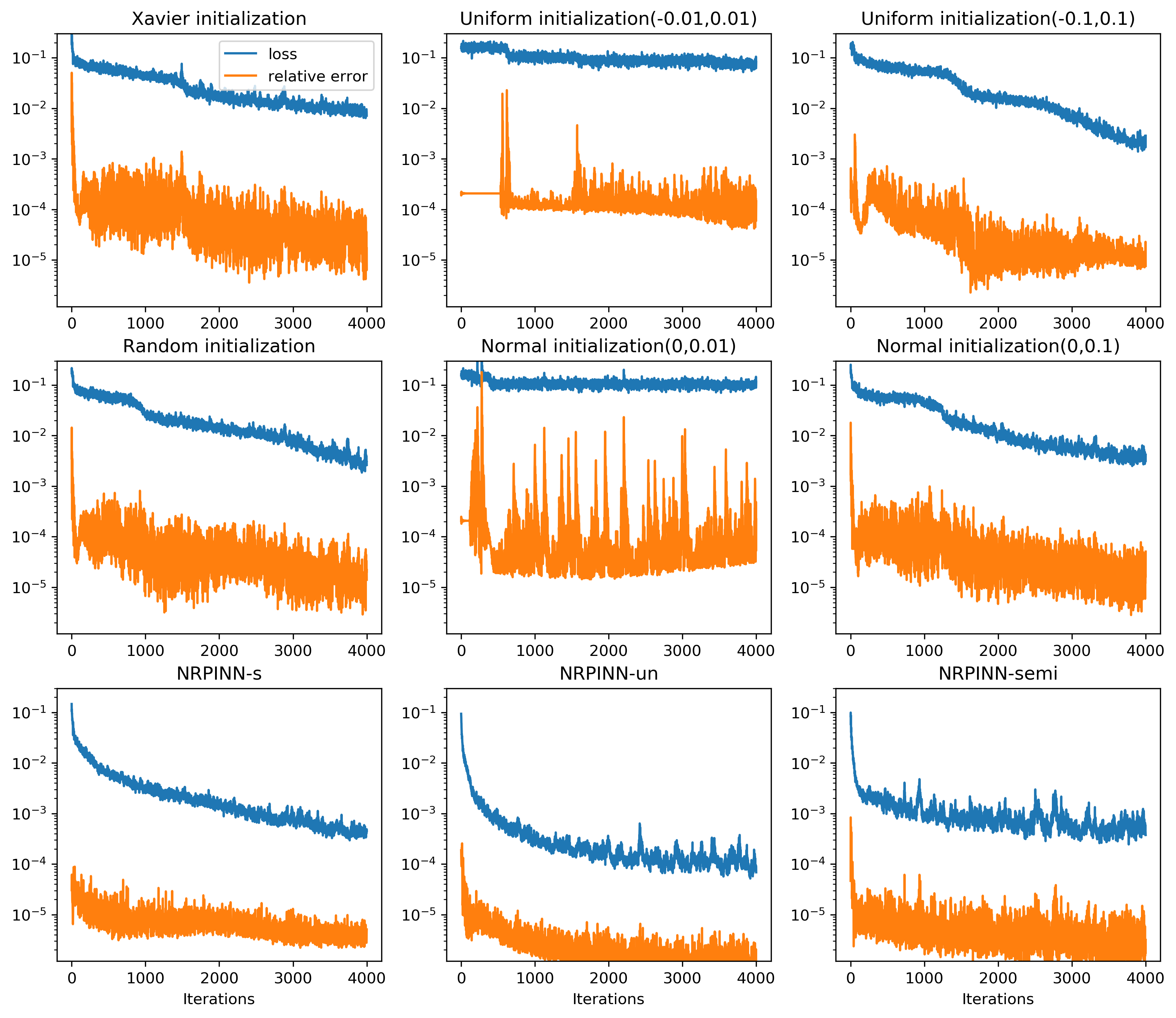}
	\caption{Two-dimensional Poisson equation: History and relative loss vs. epochs under different initialization methods. The first row presents Xavier initialization and two uniform distribution initialization with $U(-0.01,0.01)$ and $U(-0.01,0.1)$. The second row presents random and normal distribution initialization with $\mathcal{N}\left(0,0.01\right)$ and $\mathcal{N}\left(0,0.1\right)$. The last row presents the historical loss by the NRPINN-s, NRPINN-un, and NRPINN-semi over the PINN with Xavier initialization, respectively.}
	\label{diff_ini_poisson_2d}
\end{figure}

Fig.\ref{8} shows that the result of the PINN (the first row) with Xavier initialization and NRPINN-s, NRPINN-un, and NRPINN-semi (the last three rows). 
Obviously, even if PDEs with different heat sources are used as the tasks for NRPINN, the accuracy of NRPINN-s, NRPINN-un, and NRPINN-semi is visibly improved and the loss is clearly decreasing much fast. 
The training by NRPINN-un and NRPINN-semi has converged till 200 iteration. 
In the loss column, the predicted loss (green) denotes the relative $L_{2}$ error, where the predicted loss of NRPINN-un is lower 100 times over the PINN with Xavier initialization. 
The PDE loss as well as the BC loss of NRPINN-s, NRPINN-un, and NRPINN-semi converge faster than PINNs with the Xavier initialization. 
To analyze the effect of different initialization methods, Fig.\ref{diff_ini_poisson_2d} shows the history and relative loss with different initializations, where NRPINN-s, NRPINN-un, and NRPINN-semi outperform the PINN with other initialization methods. 
Especially, the relative error by the NRPINN-un reaches $1e-06$, whose performance is best. 
In this case, Xavier initialization has similar to the random distribution initialization. 

\begin{table}[!h]
	\caption{\label{poisson_2d_rel} Two-dimensional Poisson equation: The MAEs of prediction solution with different initialization methods after 4,000 iterations. The best result of NRPINN is highlighted.}
	\begin{tabular}{l|l}
		\hline
		\multicolumn{1}{c|}{Initialization method} & \multicolumn{1}{c}{MAE} \\ \hline
		Xavier                                     & 1.6974036e-04           \\ \hline
		Uniform(-0.01,0.01)                        & 1.4415e-04              \\ \hline
		Uniform(-0.1,0.1)                          & 7.58242231e-05          \\ \hline
		Random                                     & 1.40351876e-05          \\ \hline
		Normal(0,0.01)                             & 2.2043e-04              \\ \hline
		Normal(0,0.1)                              & 5.07256991e-05          \\ \hline
		NRPINN-s                                    & 5.08864878e-06          \\ \hline
		\textbf{NRPINN-un}                          & \textbf{6.44704471e-07} \\ \hline
		NRPINN-semi                                 & 3.15731108e-06          \\ \hline
	\end{tabular}
\end{table}

Table \ref{poisson_2d_rel} shows the MAEs between prediction solutions and exact solutions for different initialization methods after 4,000 iterations. From the table, we can find that NRPINN achieves higher accuracy than the PINN with other initialization methods. In this case, the performance of the PINN with normal distribution initialization is better than Xavier initialization. However, the MAE of the NRPINN-un tends to be $6.4470e-07$, which is significantly better than $5.0725e-05$ by the PINN with normal distribution initialization. In addition, the performances of NRPINN-s and NRPINN-semi are better than the PINN with normal distribution initialization.
\begin{figure}[!h]
	\includegraphics[scale=0.28]{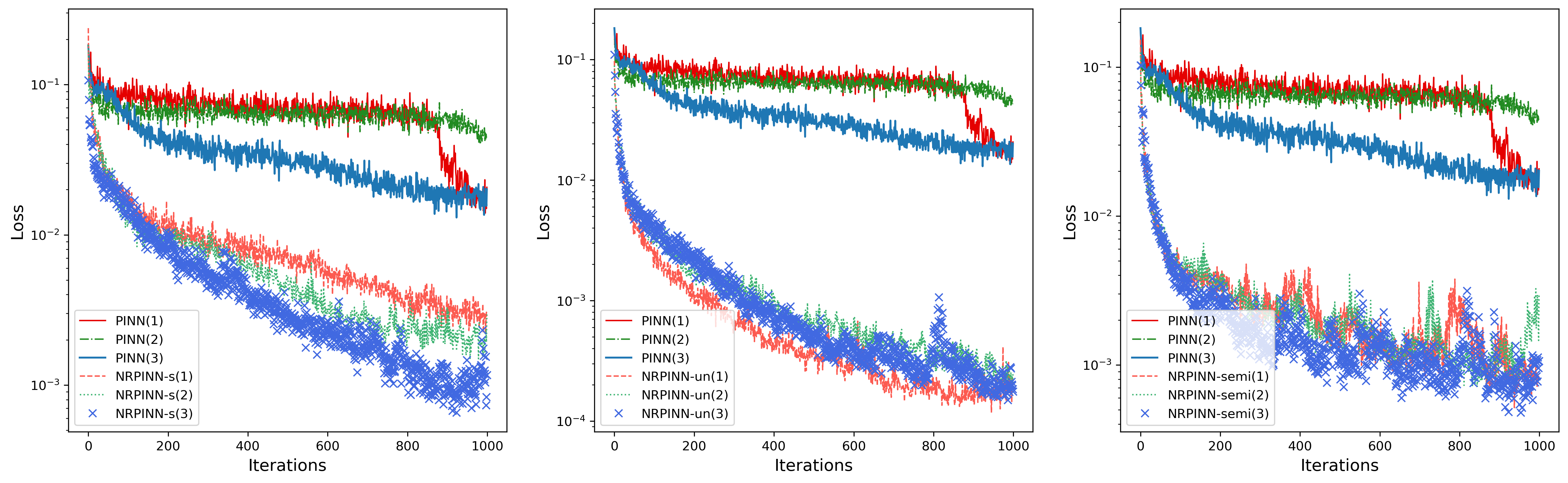}
	\caption{Two dimensional Poisson equation: Loss vs. epochs for three different locations of heat sources. The figures (from left to right) represent the historical loss by the NRPINN-s, NRPINN-un, and NRPINN-semi over the PINN with Xavier initialization.}
	\label{different_loss}
\end{figure}
To further analyze the acceleration effect of the NRPINN for solving Eq.(\ref{source_n}), the position of eight heat sources in the Fig.\ref{8h} (left) is changed to obtain three PDEs. 
For illustration, we use NRPINN-s, NRPINN-un, and NRPINN-semi to solve another three PDEs and the experimental results shown in Fig.\ref{different_loss}. As Fig.\ref{different_loss} shows, the training losses by NRPINN-s, NRPINN-un, and NRPINN-semi for the three PDEs are decreasing faster than the PINN with Xavier initialization. 
More apparently, the NRPINN can accelerate the training of any PDEs in the Eq.(\ref{schem}) and achieves similar acceleration effects for different PDEs.
So one can see that NRPINN can efficiently solve a class of similar tasks.

\subsection{Burgers equation}
\label{Burger equation}

\begin{figure}[!h]
	\includegraphics[scale=0.4]{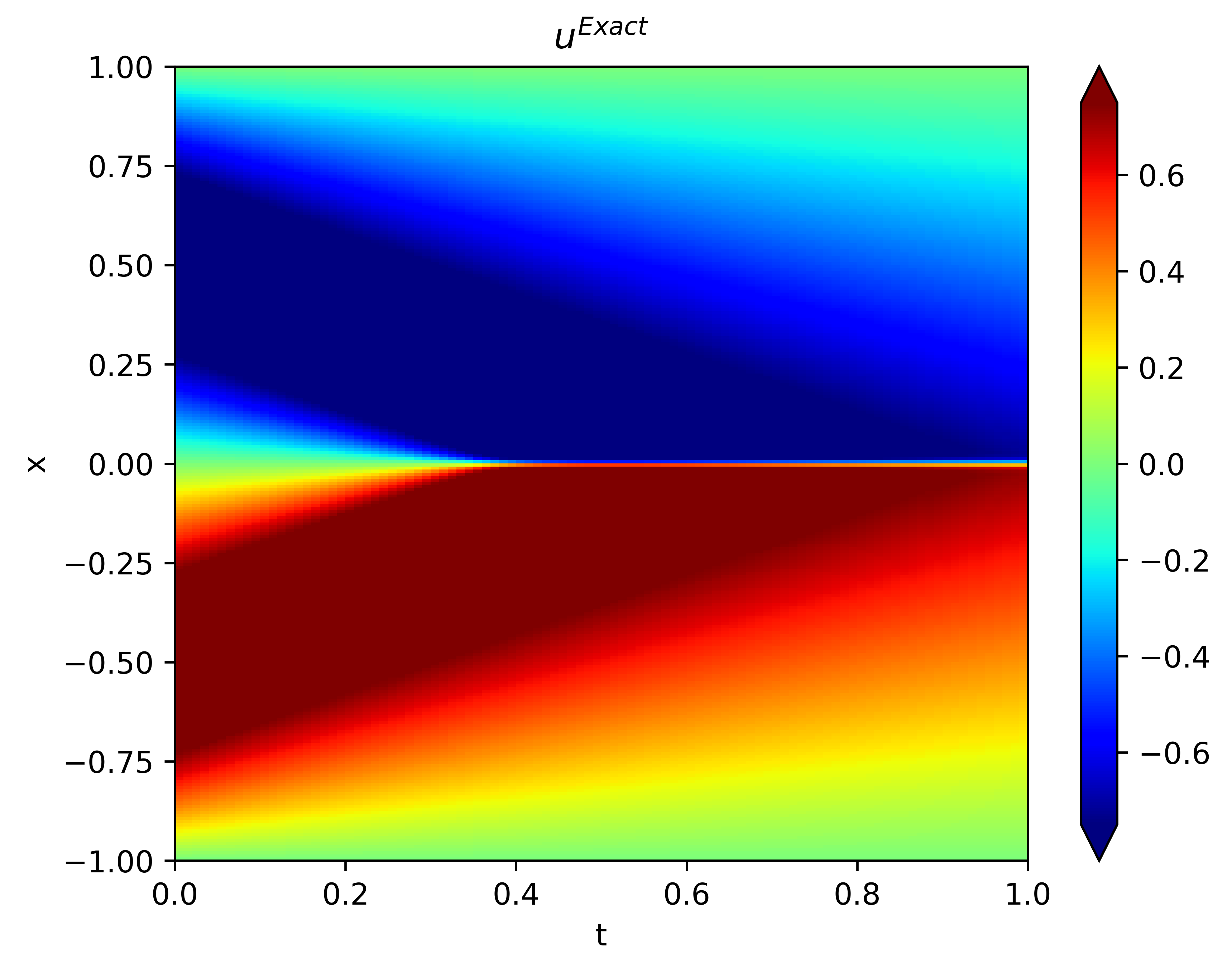}
	\caption{The solution of Burgers equation.}
	\label{burgers}
\end{figure}

Before solving the Burgers equation, the high-order information is given by
\begin{equation}
\label{burger_high}
\begin{aligned}
u_{t}+u u_{x} &=v u_{x x}, x \in[-1,1], t>0, \\
u(x, 0) &=-\sin (\pi x), \\
u(-1, t) &=u(1, t)=0,
\end{aligned}
\end{equation}
with viscous parameter $v \in[0,0.1 / \pi]$. 
Each task obtained from the high-order information represents a governing equation in Eq.(\ref{burger_high}).
The zero-order information is the set of numerical solutions under different $v$. Each task obtained from the zero-order information represents the numerical solution under one value $v \in [0.005 / \pi, 0.1 / \pi]$.
In this case, the Burgers equation to be solved is when $v=0.01 / \pi$. 
The analytical solution can be obtained using the Hopf–Cole transformation and its graph is shown in Fig.\ref{burgers}, see Basdevant et al. \cite{basdevant1986spectral} for more details. 

In this case, the number of hidden layers is eight and in each layer 20 neurons are used. 
In the new Reptile initialization part, we consider three types of new Reptile initialization: 
(i) for supervised learning, we assume that 20 tasks can be obtained from the zero-order information. Then, we sample 1,000 training points under each task, confirm loss function $\mathcal{L}_{z}(\theta)$, and Adam is used to update the parameter $\theta$ 10,000 times with the learning rate of 0.001. 
(ii) for unsupervised learning, we randomly select 50 tasks, sample 1,000 residual points and 1,000 training points on the boundary, and update the parameters of the NN 10,000 times. 
(iii) for semi-supervised learning, we select 75 tasks, of which half is used for supervised learning and the other half for unsupervised learning. 
In the part of solving PDEs by PINNs, the PINN with initialization parameters is used to solve the PDEs. The number of residual training points is 10,000, the number of training data points on the boundary is 5,000, and the number of true data is 0.

\begin{figure}[!h]
	\centering
	\includegraphics[scale=0.20]{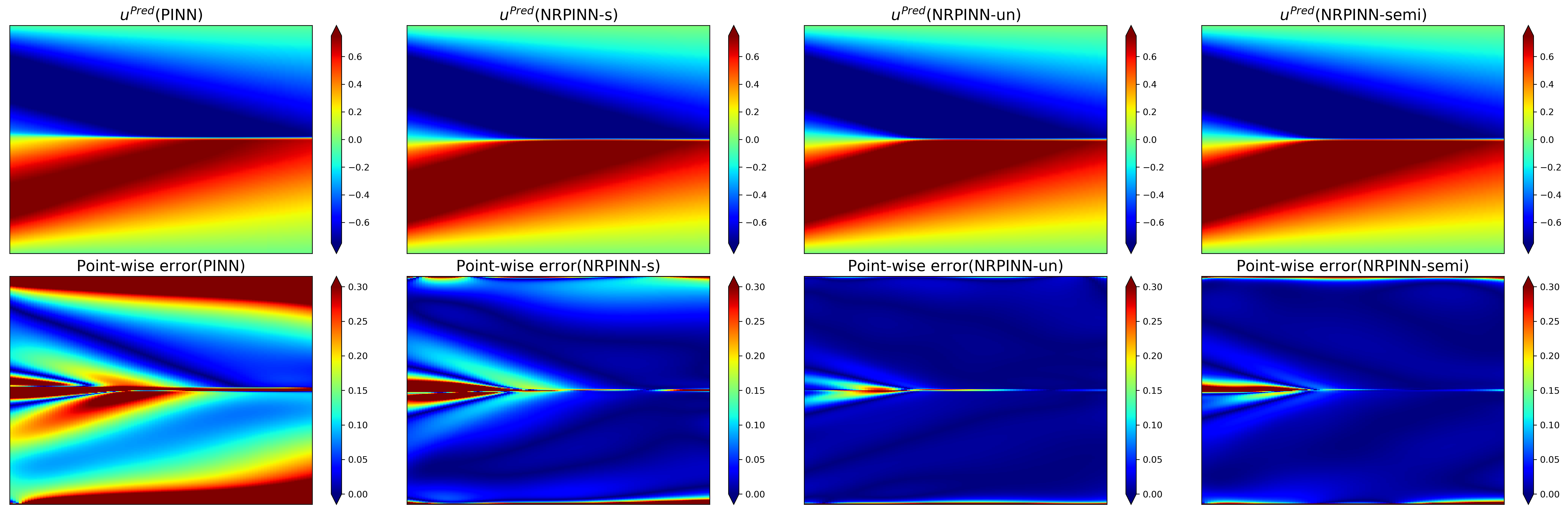}
	\caption{Burgers equation: Comparison of the predicted solution (the first row) as well as point-wise error (the last row) by PINNs with Xavier initialization, the NRPINN-s, NRPINN-un and NRPINN-semi after $2,000$ iterations.}
	\label{burgers_u}
\end{figure}

\begin{figure}[!h]
	\includegraphics[scale=0.38]{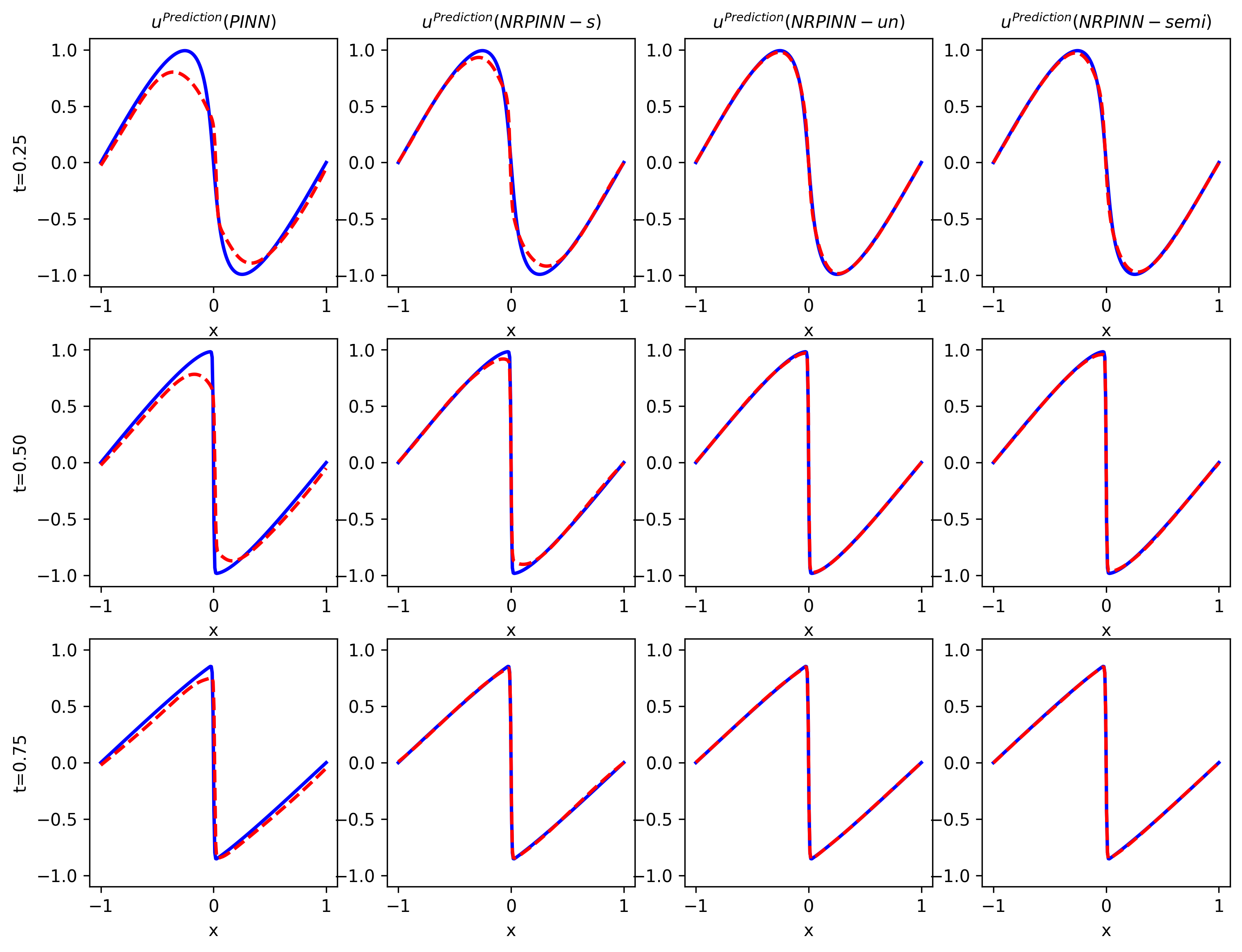}
	\caption{Burgers equation: Comparison of the exact solution with the solution given by PINNs (the first column) with Xavier initialization as well as the NRPINN (the last three columns) with supervised, unsupervised, and semi-supervised learning obtained after 2,000 iterations.}
	\label{burgers_time}
\end{figure}

\begin{figure}[!h]
	\includegraphics[scale=0.34]{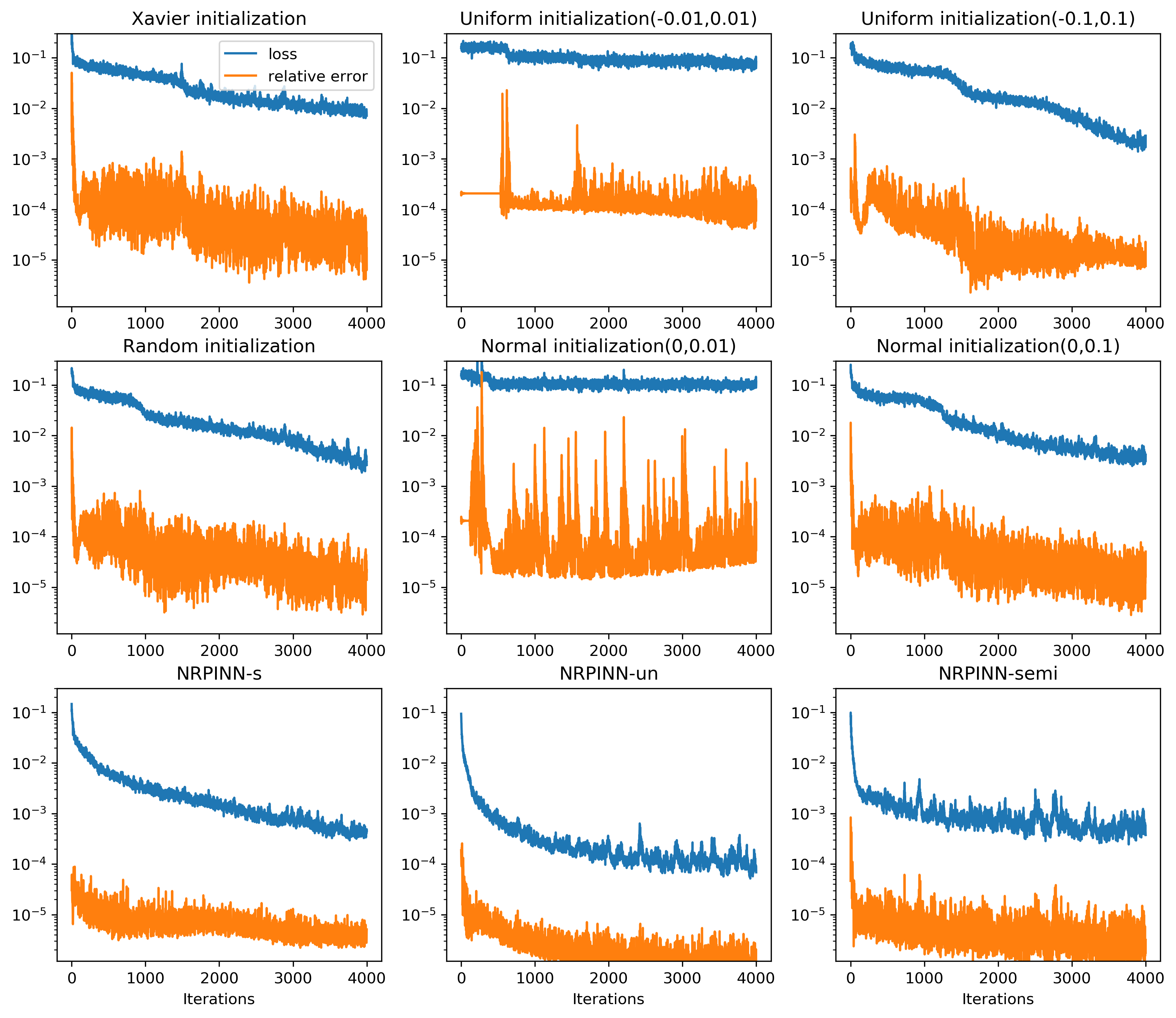}
	\caption{Burgers equation: History and relative loss vs. epochs under different initialization. The first row presents Xavier as well as two uniform distribution initialization with $U(-0.01,0.01)$ and $U(-0.01,0.1)$. The second row presents random and normal distribution initialization with $\mathcal{N}\left(0,0.01\right)$ and $\mathcal{N}\left(0,0.1\right)$. The last row presents the NRPINN-s, NRPINN-un, and NRPINN-semi, respectively.}
	\label{burgers_loss}
\end{figure}

Fig.\ref{burgers_u} shows the predicted solution (top row) as well as the point-wise error (bottom row) by PINNs with Xavier initialization, NRPINN-s, NRPINN-un, and NRPINN-semi. 
NRPINN-s, NRPINN-un, and NRPINN-semi accurately capture all the dispersive waves in the solution and the point-wise errors in the entire domain are very low. In contrast, the PINN with Xavier initialization fails to predict the solution accuracy and its point-wise error is very high compared to NRPINN-s, NRPINN-un, and NRPINN-semi. 
Fig.\ref{burgers_time} gives the comparison of solutions by the PINN with Xavier initialization, NRPINN-s, NRPINN-un, and NRPINN-semi at various t locations. 
In the left column, Xavier initialization is used for the PINN and the right three columns present the results of NRPINN-s, NRPINN-un, and NRPINN-semi, where one can see that the accuracy of NRPINN-s, NRPINN-un, and NRPINN-semi over the PINN is significantly improved. 
Fig.\ref{burgers_loss} shows the loss history and relative $L_{2}$ error over the number of iterations with different initializations methods. The loss and relative $L_{2}$ error of the NRPINN over the PINN with other initialization methods are visibly decreasing faster. 
The PINN with Xavier initialization outperforms other initializations methods but its loss and relative error reach $1e-02$, far less than the NRPINN-un.

Table \ref{burgers_rel} lists the MAEs between prediction solutions and exact solutions for different initialization methods after 2,000 iterations. The MAEs of NRPINN-s, NRPINN-un, and NRPINN-semi are far less than the PINN with other initialization methods. The performance of NRPINN-un is the best and the MAE of NRPINN-un tends to be $8.582e-05$. The MAE of the PINN with uniform distribution initialization tends to be $3.774e-01$, which is worse than NRPINN-s, NRPINN-un and NRPINN-semi.

\begin{table}[!b]
	\caption{\label{burgers_rel} Burgers equation: The MAEs of
		prediction solution with different initialization methods after 2,000 iterations. The best result of NRPINN is highlighted.}
	\begin{tabular}{l|l}
		\hline
		\multicolumn{1}{c|}{Initialization method} & \multicolumn{1}{c}{MAE} \\ \hline
		Xavier                                     & 8.871e-03               \\ \hline
		Uniform(-0.01,0.01)                        & 3.774e-01               \\ \hline
		Uniform(-0.1,0.1)                          & 7.656e-02               \\ \hline
		Random                                     & 2.578e-02               \\ \hline
		Normal(0,0.01)                             & 5.6194e-02              \\ \hline
		Normal(0,0.1)                              & 3.3010e-02              \\ \hline
		NRPINN-s                                    & 1.3842e-03              \\ \hline
		\textbf{NRPINN-un}                          & \textbf{8.582e-05}      \\ \hline
		NRPINN-semi                                 & 1.7821e-04              \\ \hline
	\end{tabular}
\end{table}

\subsection{Schr\"odinger equation}\label{sec34}
\begin{figure}[!h]
	\includegraphics[scale=0.3]{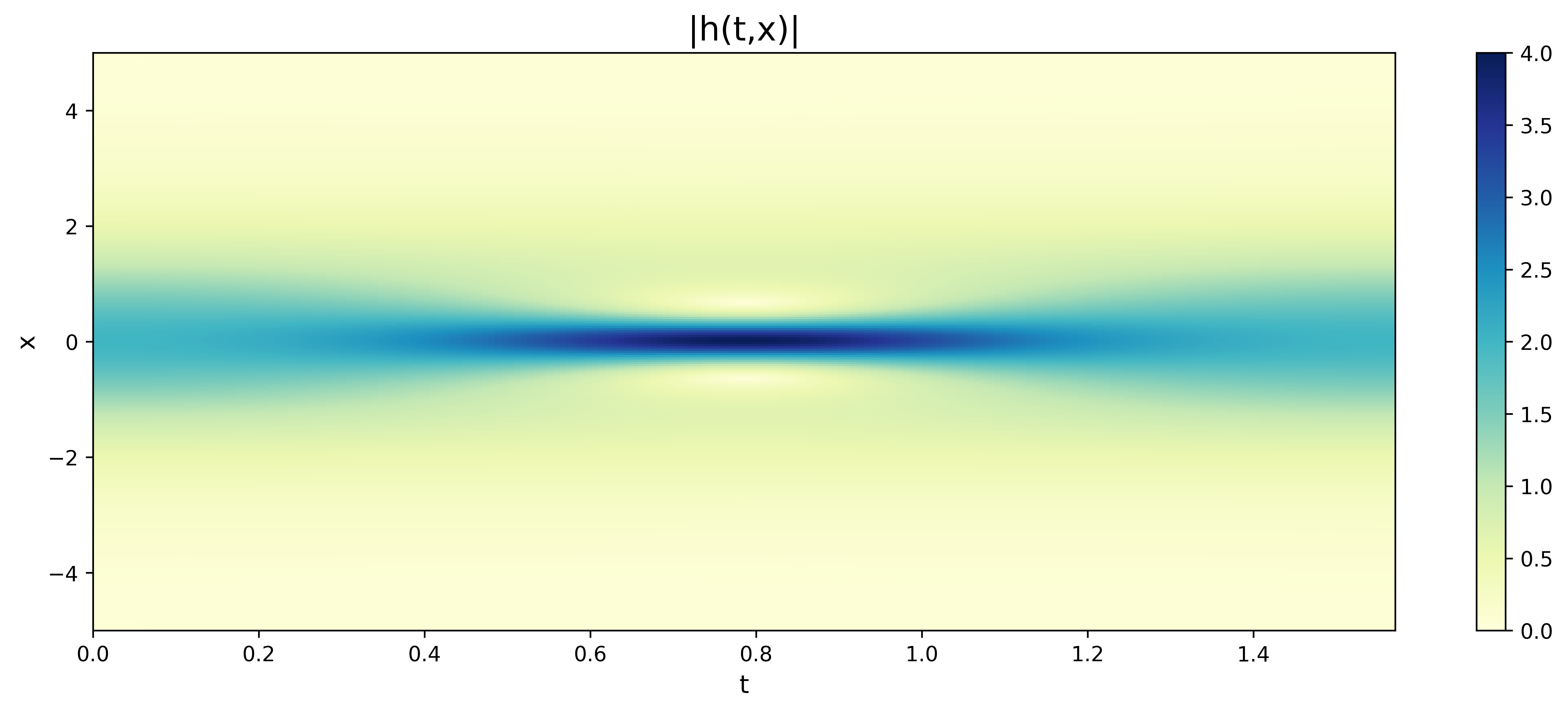}
	\caption{The exact solution of the Schrödinger equation.}
	\label{Sch}
\end{figure}

Before solving the Schr\"odinger equation, the high-order information is given by
\begin{equation}
\label{sch}
\begin{array}{l}
i h_{t}+\lambda h_{x x}+|h|^{2} h=0, \\
h(t,-5)=h(t, 5), \\
h_{x}(t,-5)=h_{x}(t, 5),\\
h(0,x)=2sech(x),
\end{array}
\end{equation}
where $x \in[-5,5]$, $t \in[0, \pi / 2]$ and $\lambda\sim \text{uniform}[0,1]$. 
Each task obtained from the high-order information represents a governing equation in Eq.(\ref{sch}). 
In this case, the Schr\"odinger equation to be solved is the PDE when $\lambda = 0.5$ and its graph is shown in Fig.\ref{Sch}. 
Considering the Eq.\ref{sch}, we consider unsupervised learning for the new Reptile initialization. 
We use four hidden layers with 100 neurons in each layer. 
In the part of new Reptile initialization, we randomly select 100 tasks for unsupervised learning, sample 1,000 residual points and 1,000 training points on the boundary, and update the parameters $\theta$ 10,000 times. In the part of solving PDEs by PINNs, we consider 20,000 residual training points, 1,000 training data points on the boundary, and 0 real training data.

\begin{figure}[!h]
	\includegraphics[scale=0.039]{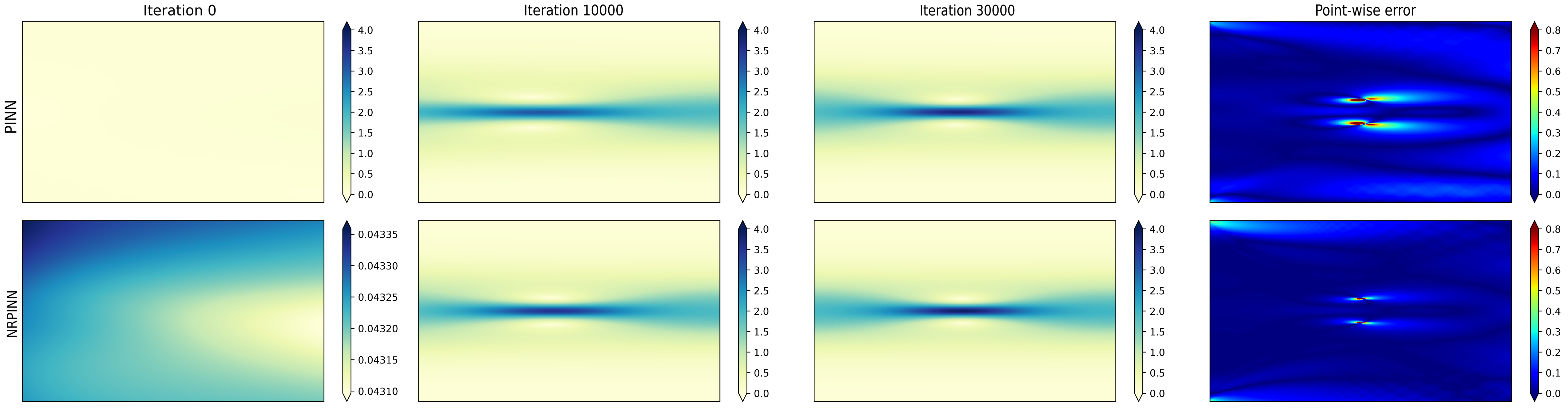}
	\caption{Schr\"odinger equation: Top row is the results of PINNs and bottom row is the results of the NRPINN. From left to right, the predicted solution at 0, 10,000 and 30,000 iterations, and point-wise error at 30,000 iterations.}
	\label{schrodinger}
\end{figure}

\begin{figure}[!h]
	\includegraphics[scale=0.45]{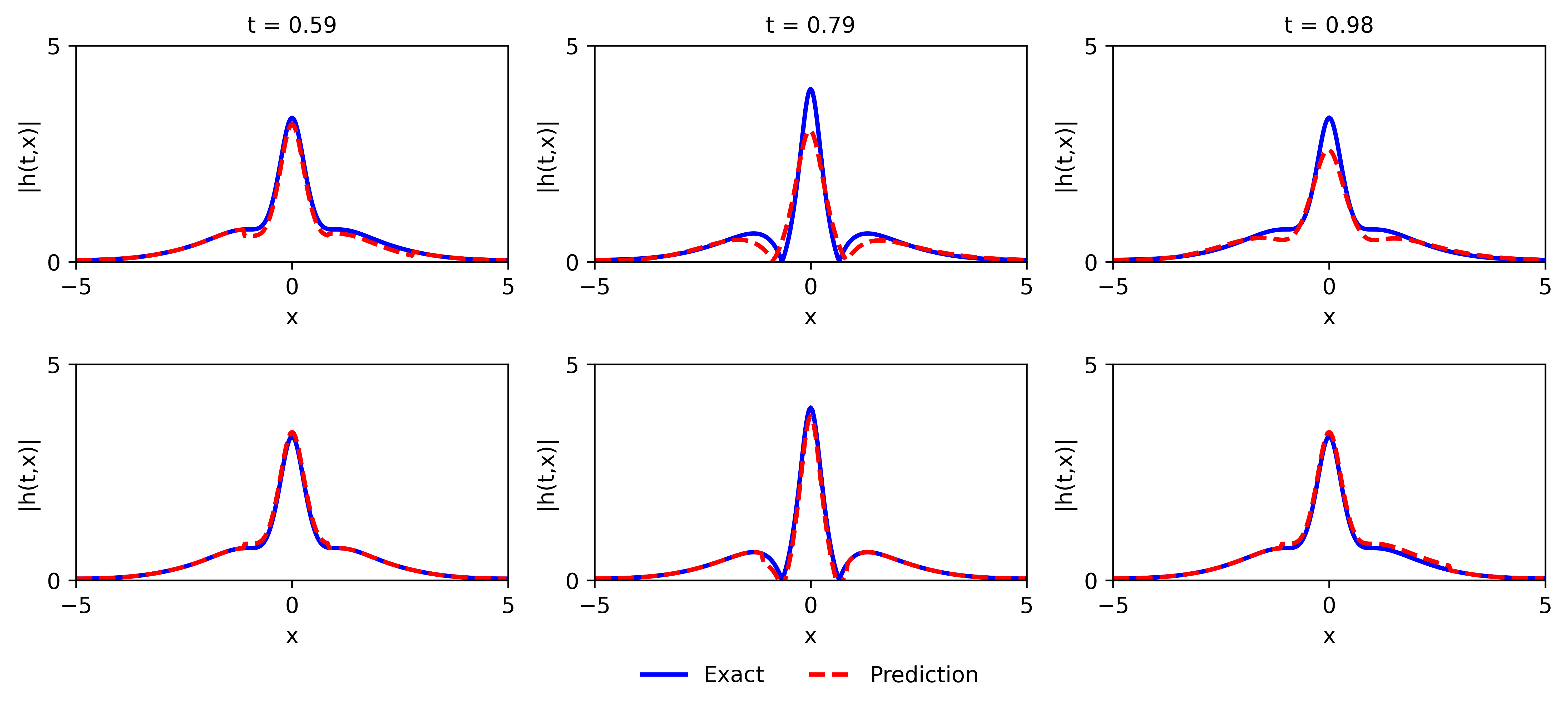}
	\caption{Schr\"odinger equation: Comparison of the exact solution with the solution given by PINNs with Xavier initialization (top) and the NRPINN-un (bottom) obtained after 30,000 iterations.}
	\label{sch_time}
\end{figure}

\begin{figure}[!h]
	\includegraphics[scale=0.34]{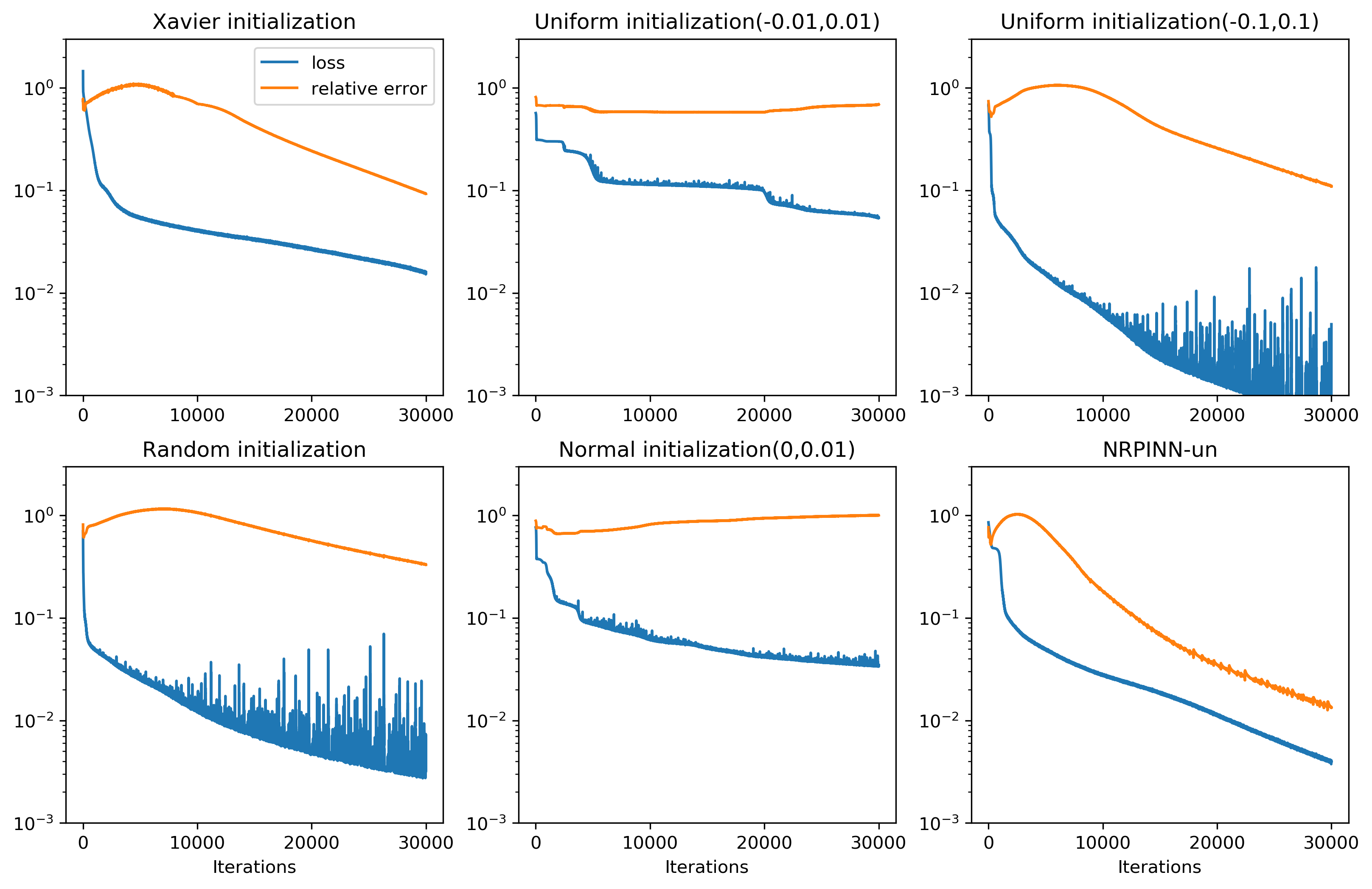}
	\caption{Schr\"odinger equation: History and relative loss vs. epochs under different initialization methods. The first row presents Xavier initialization and two uniform distribution initialization with $U(-0.01,0.01)$ and $U(-0.01,0.1)$. The second row presents the PINN with random and normal distribution initialization($\mathcal{N}\left(0,0.01\right)$), and the NRPINN-un, respectively.}
	\label{sch_loss}
\end{figure}

Fig.\ref{schrodinger} shows the predicted solution at various iterations, and the point-wise error at 30,000 iterations by the PINN (top row) with Xavier initialization and NRPINN-un (bottom row). 
Compared to the PINN, NRPINN-un accurately captures the feature of the solution faster and the point-wise error at 30,000 iterations in the entire domain is lower. 
Fig.\ref{sch_time} gives the comparison of solution by the PINN and the NRPINN-un at $t=0.25, 0.5, 0.75$. 
The top figure shows results of the PINN with Xavier initialization and the bottom figure presents results of NRPINN-un, where the accuracy of NRPINN-un over PINNs is visibly improved. 
Especially, the predicted solution at $t=0.79$ of PINNs is worst than NRPINN-un, which accords with points-wise error in Fig.\ref{sch_time}. 
Fig.\ref{sch_loss} shows the loss history and the relative $L_{2}$ error over the number of iterations for different initialization methods, where the loss and the relative $L_{2}$ error of NRPINN-un is decreasing faster over other initialization methods.

Table \ref{sch_rel} lists the MAEs between prediction solutions and exact solutions for different initialization methods under 30,000 iterations. The MAE of the NRPINN-un tends to be $1.3406e-02$, which is better than $1.0742e-01$ by the PINN with Xavier initialization. The PINN with uniform distribution initialization tends to be $1.0060$, which is far worse than $1.3406e-02$ by Xavier initialization method. In summary, the NRPINN endows the PINN with a good start to achieve higher accuracy.

\begin{table}[!h]
	\caption{\label{sch_rel} Schr\"odinger equation: The MAEs of prediction solution with different initialization methods after 30,000 iterations. The best result of NRPINN is highlighted.}
	\begin{tabular}{l|l}
		\hline
		\multicolumn{1}{c|}{Initialization method} & \multicolumn{1}{c}{MAE} \\ \hline
		Xavier                                     & 1.0742e-01              \\ \hline
		Uniform(-0.01,0.01)                        & 6.9054e-01              \\ \hline
		Uniform(-0.1,0.1)                          & 1.0934e-01              \\ \hline
		Random                                     & 3.3117e-01              \\ \hline
		Normal(0,0.01)                             & 1.0060                  \\ \hline
		\textbf{NRPINN-un}                          & \textbf{1.3406e-02}     \\ \hline
	\end{tabular}
\end{table}

\subsection{Inverse problem for Burgers}

\label{4}
In response to the lack of data, in this section, we consider quickly identifying the unknown parameters in the PDEs with a small amount of real data by NRPINN. The parameterized viscous Burgers equation is given by
\begin{equation}
\label{inver_burger}
u_{t}+\mathcal{N}[u ; v]=0, x \in \Omega \subset \mathbb{R}, v \in[0,0.1/ \pi], t>0,
\end{equation}
where $u(t,x)$ is hidden solution and $\mathcal{N}[u ; v]=u^{2} / 2-v u_{x}$ is a parameterized nonlinear term. 
In this case, we consider solving the PDE when viscosity coefficient $v$is equal to $0.01/\pi$. Eq.\ref{inver_burger} can be considered as the high-order information for the new Reptile initialization. The zero-order information is the set of the solutions for different $v$. 
The detailed experimental parameters setting for supervised, unsupervised, and semi-supervised learning are the same as section \ref{Burger equation}. 
\begin{figure}[!h]
	\includegraphics[scale=0.3]{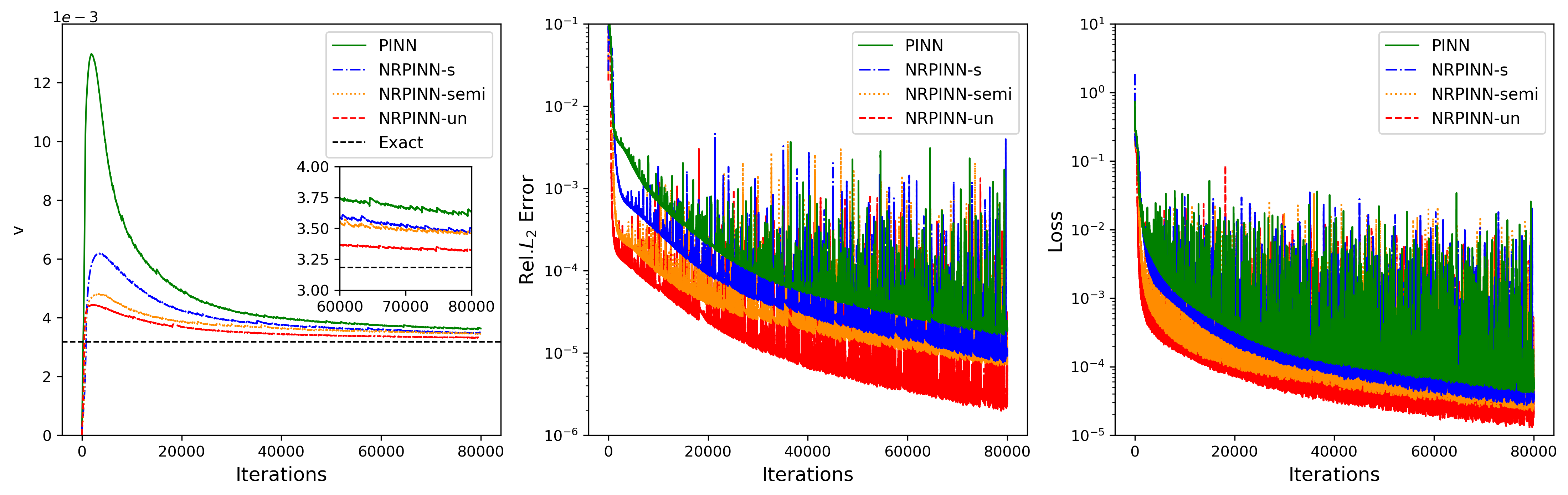}
	\caption{Inverse problem: From left to right, $v$ history, $L_{2}$ error, and MSE history over number of iterations of iterations for the PINN with Xavier initialization, NRPINN-s, NRPINN-un, and NRPINN-semi, respectively.}
	\label{nu_inverse}
\end{figure}

\begin{figure}[!h]
	\includegraphics[scale=0.3]{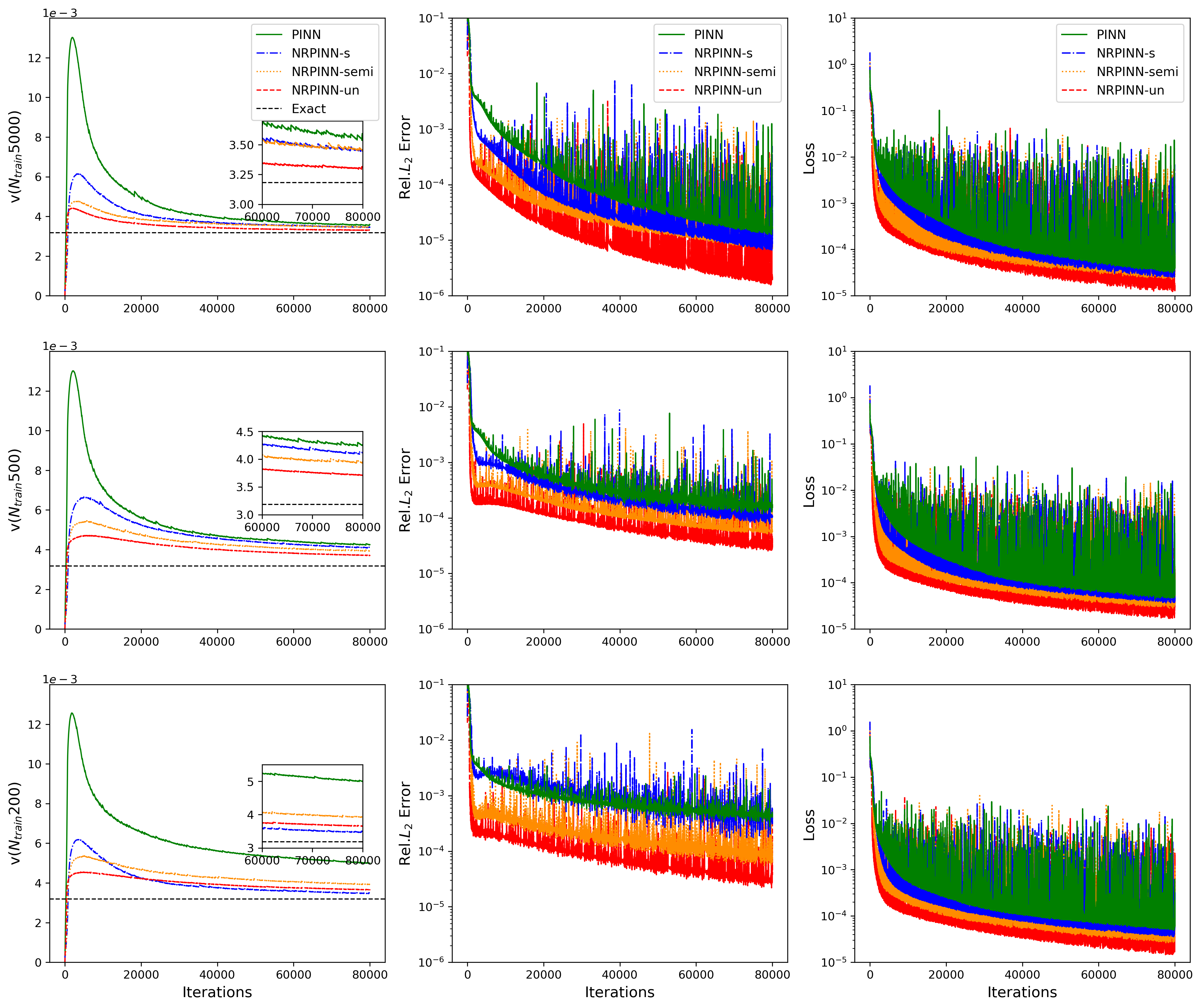}
	\caption{Inverse problem: From left to right, $v$ history, $L_{2}$ error, and MSE history over number of iterations for the PINN, NRPINN-s, NRPINN-un, and NRPINN-semi with 10,000 $N_{train}$. }
	\label{diff_data}
\end{figure}

Fig.\ref{nu_inverse} (left to right) shows the $v$ history, $L_{2}$ error, and loss versus number of iterations, respectively. 
In the left figure, the predicted $v$ by NRPINN-s, NRPINN-un, and NRPINN-semi visibly converge faster towards its exact value over the PINN. 
After 80,000 iterations, the predicted $v$ by NRPINN-s, NRPINN-un, and NRPINN-semi are $0.0034965$, $0.0033251$, and $0.0034558$, respectively. The relative errors by NRPINN-s, NRPINN-un, and NRPINN-semi are $9.85\%$, $4.46\%$, and $8.57\%$, whereas the predicted $v$ by the PINN with Xavier initialization is $0.0036330$ and the relative error is $14.13\%$. 
To analyze the dependence by the PINN with Xavier initialization and NRPINN on real training data ($N_{train}$), we use different number of real data.
Fig.\ref{diff_data} shows the $v$ history, $L_{2}$ error, and loss versus number of iterations with different real training data, where NRPINN-s, NRPINN-un, and NRPINN-semi always predict unknown parameters $v$ faster and achieve higher accuracy than the PINN with Xavier initialization. 
For $N_{train}$ is equal to $200$, the predicted $v$ by the PINN is $0.004990$ and the relative error is $56.94\%$, whereas the relative error by NRPINN-s, NRPINN-un, and NRPINN-semi reaches $36.41\%$, $14.80\%$, and $23.32\%$, respectively, where one can see that NRPINN maintains good predictive performance when the amount of real training data is small.

\begin{figure}[!h]
	\includegraphics[scale=0.32]{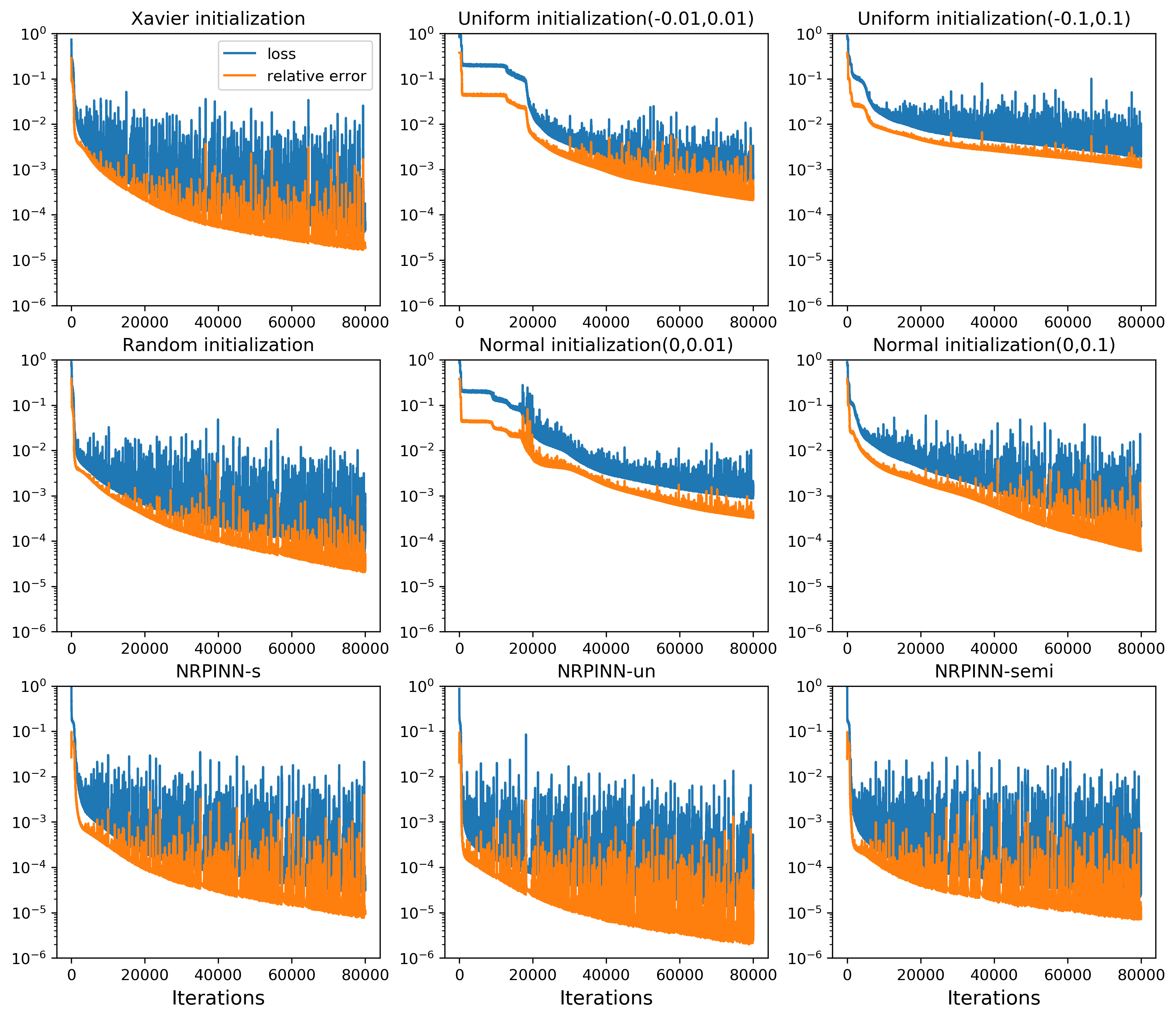}
	\caption{Inverse problem: History and relative loss vs. epochs under different initialization methods. The first row presents Xavier initialization and two uniform distribution initialization with $U(-0.01,0.01)$ and $U(-0.01,0.1)$. The second row presents random and normal distribution initialization with $\mathcal{N}\left(0,0.01\right)$ and $\mathcal{N}\left(0,0.1\right)$. The last row presents NRPINN-s, NRPINN-un, and NRPINN-semi, respectively.
	}
	\label{nu_diff_data}
\end{figure}

\begin{figure}[!h]
	\includegraphics[scale=0.27]{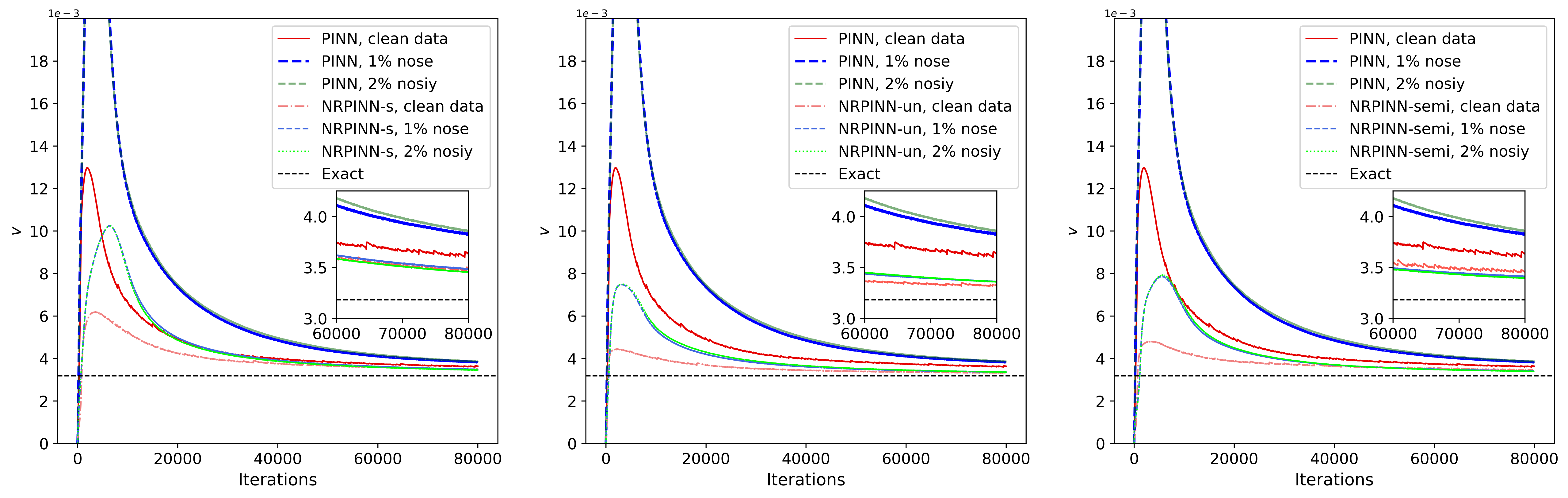}
	\caption{Inverse problem: The $v$ history with $10,000$ clean data, $1\%$ or $2\%$ noise data under 80,000 iterations. The figures (from left to right) represent the comparison result by NRPINN-s, NRPINN-un, and NRPINN-semi over PINNs with Xavier initialization, respectively.}
	\label{nu_bias}
\end{figure}

\begin{figure}[!h]
	\includegraphics[scale=0.25]{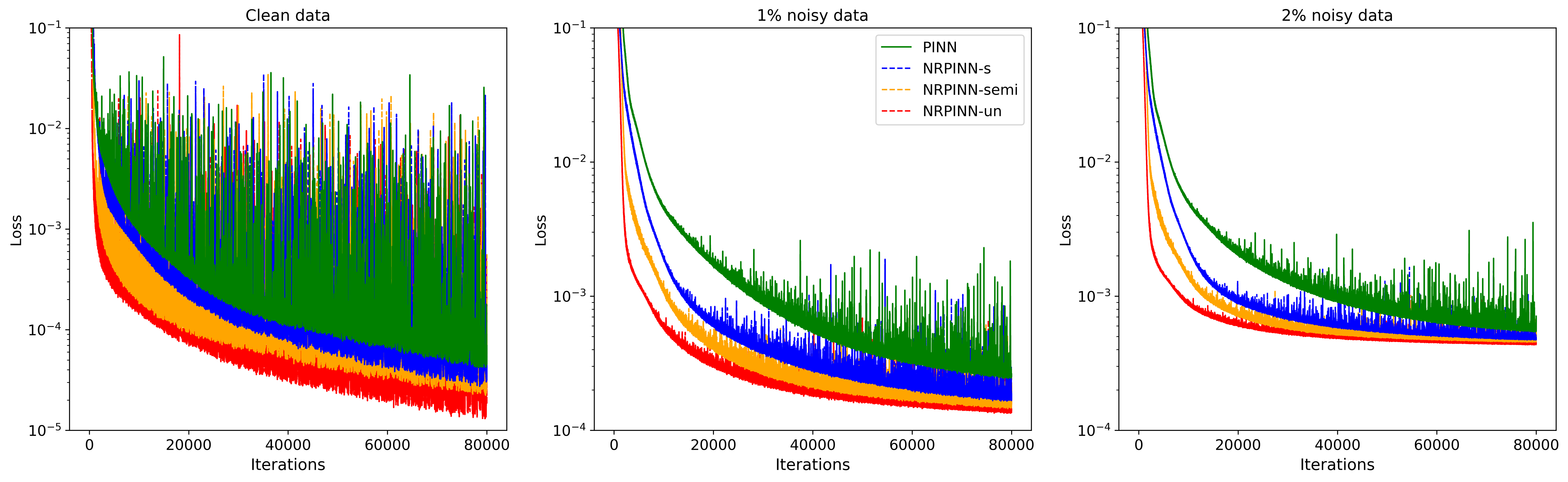}
	\caption{Inverse problem: The historical loss vs epoch with 10,000 noisy $N_{data}$, $1\%$, and $2\%$ noise data. The figures (left to right) present the historical loss by the NRPINN-s, NRPINN-un, and NRPINN-semi over PINNs with Xavier initialization.}
	\label{loss_bias}
\end{figure}

\begin{table}[!h]
		\caption{\label{noisy data} The relative error of $v$ with different number of clean data, $1\%$ and $2\%$ noise data under 80,000 iterations.}
	\begin{tabular}{c|c|c|c|c|c|cccccll}
		\cline{1-7}
		$N_{data}$  & \multicolumn{3}{c|}{200}                      & \multicolumn{3}{c}{500}                               & \multicolumn{3}{c}{}                                               & \multicolumn{3}{c}{}                                                        \\ \cline{1-7}
		Noise       & clean            & 1\%              & 2\%     & clean            & 1\%              & 2\%              &                      &                      &                      &                      & \multicolumn{1}{c}{} & \multicolumn{1}{c}{}          \\ \cline{1-7}
		PINN        & 56.94\%          & \textbf{69.74\%} & 75.42\% & 33.57\%          & \textbf{19.36\%} & 23.12\%          & \textbf{}            &                      &                      & \textbf{}            & \multicolumn{1}{c}{} & \multicolumn{1}{c}{} \\ \cline{1-7}
		NRPINN-s    & 36.41\%          & \textbf{14.9\%1} & 45.63\% & 29.19\%          & 13.27\%          & 7.49\%           &                      &                      &                      &                      & \multicolumn{1}{c}{} & \multicolumn{1}{c}{}          \\ \cline{1-7}
		NRPINN-un   & 14.80\%          & 17.43\%          & 22.93\% & 16.53\%          & \textbf{10.99\%} & 18.98\%          & \textbf{}            & \textbf{}            &                      & \textbf{}            & \multicolumn{1}{c}{} & \multicolumn{1}{c}{} \\ \cline{1-7}
		NRPINN-semi & 23.32\%          & 12.38\%          & 22.77\% & 24.11\%          & 19.77\%          & 18.10\%          &                      &                      &                      &                      & \multicolumn{1}{c}{} & \multicolumn{1}{c}{}          \\ \cline{1-7}
		$N_{data}$  & \multicolumn{3}{c|}{5000}                     & \multicolumn{3}{c}{10000}                             &                      &                      &                      &                      &                      &                               \\ \cline{1-7}
		Noise       & clean            & 1\%              & 2\%     & clean            & 1\%              & 2\%              & \multicolumn{1}{l}{} & \multicolumn{1}{l}{} & \multicolumn{1}{l}{} & \multicolumn{1}{l}{} &                      &                               \\ \cline{1-7}
		PINN        & \textbf{12.79\%} & \textbf{23.59\%}          & 21.27\% & \textbf{14.13\%} & 20.26\%          & \textbf{21.15\%} & \multicolumn{1}{l}{} & \multicolumn{1}{l}{} & \multicolumn{1}{l}{} & \multicolumn{1}{l}{} &                      &                               \\ \cline{1-7}
		NRPINN-s    & 8.62\%           & 10.55\%          & 10.32\% & 9.85\%           & 9.44\%           & 8.55\%           & \multicolumn{1}{l}{} & \multicolumn{1}{l}{} & \multicolumn{1}{l}{} & \multicolumn{1}{l}{} &                      &                               \\ \cline{1-7}
		NRPINN-un   & \textbf{4.12\%}  & \textbf{6.97\%}  & 5.23\%  & \textbf{4.46\%}  & 5.59\%           & \textbf{5.51\%}  & \multicolumn{1}{l}{} & \multicolumn{1}{l}{} & \multicolumn{1}{l}{} & \multicolumn{1}{l}{} &                      &                               \\ \cline{1-7}
		NRPINN-semi & 8.40\%           & 6.57\%           & 1.17\%  & 8.57\%           & 7.13\%           & 6.64\%           & \multicolumn{1}{l}{} & \multicolumn{1}{l}{} & \multicolumn{1}{l}{} & \multicolumn{1}{l}{} &                      &                               \\ \cline{1-7}
	\end{tabular}
\end{table}

\begin{figure}[!h]
	\includegraphics[scale=0.28]{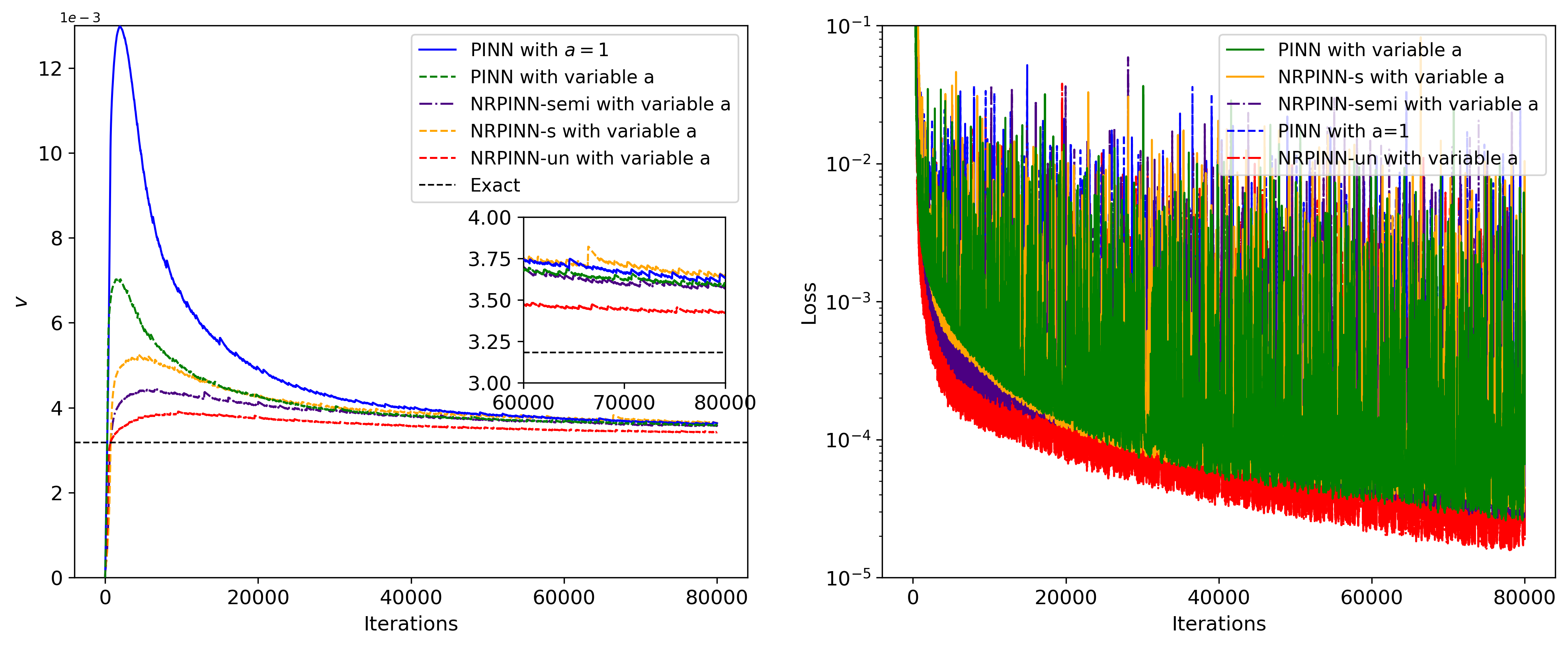}
	\caption{Inverse problem: The left figure represents $v$ history by PINNs, PINNs with adaptive activation, the NRPINN-s with adaptive activation function, the NRPINN-un with adaptive activation function and the NRPINN-semi with adaptive activation function. The right figure represents loss history corresponding to different methods}
	\label{nu_adapt}
\end{figure}

Fig.\ref{nu_diff_data} shows the history and relative loss with different initializations methods. The NRPINN has a better performance, which outperforms other initialization methods. Especially, the historical loss and relative error by NRPINN-un reach $1e-06$. 
In this case, the variation trend of Xavier initialization has similar to random distribution initialization. Fig.\ref{nu_bias} shows the $v$ history with
clean data, $1\%$, and $2\%$ noise data under 80,000 iterations. Whatever $1\%$ or $2\%$ noise data, NRPINN-s, NRPINN-un, and NRPINN-semi outperform the PINN with clean data, whose predicted $v$ is $0.0034554$, $0.0033584$ and $0.003395$. 
Faced $1\%$ and $2\%$ noise data, the predicted $v$ by PINNs is $0.003828$ and $0.0038560$, far from the exact value $0.0031831$. 
Fig.\ref{nu_bias} shows the historical loss with clean and noisy data, where NRPINN-s, NRPINN-un, NRPINN-semi achieve lower and faster historical losses than the PINN with Xavier initialization. 
Table \ref{noisy data} shows the relative error of predicted $v$ under the different number of noise data, where NRPINN-s, NRPINN-un, and NRPINN-semi achieve smaller errors than the PINN with Xavier initialization when a small number of real data is available. 
The new Reptile initialization is also used for the variants of PINNs. 
To demonstrate the new Reptile can be used for the variants of PINNs, we use new Reptile initialization for PINNs with adaptive activation function proposed by D.Jagtap et al. \cite{jagtap2020adaptive}. Fig.\ref{nu_adapt} shows the predicted $v$ and historical loss, where the PINN with variable $a$ outperform the PINN with $a=1$. Apparently, the new Reptile initialization can also be used for PINNs with adaptive activation functions to accelerate training.
\section{Conclusions}
\label{sec4}
Designing a fast and accurate deep learning method is very important for solving a large and complex PDE-based physics problem. By introducing the meta-learning algorithm, we propose a new Reptile initialization based physics-informed neural network (NRPINN) to improve the training efficiency and predicted accuracy of PINNs. 
According to learning tasks in the new Reptile initialization, the NRPINN is divided into NRPINN-s, NRPINN-un and NRPINN-semi.

To support our claim, we examined our proposed algorithm for forward and inverse problems of both smooth solutions and sharp gradient solutions, including Poisson, Burgers and Schr\"odinger equations. 
In all cases, the decay of losses by NRPINN is much faster than PINNs with other initialization methods, and its corresponding mean relative errors between predicted and the exact solution by NRPINN are much lower than PINNs with other initialization methods. 
Besides, the new Reptile initialization is a general structure for PINNs, which is also used for variants of PINNs, such as PINNs with adaptive activation [24], parareal PINNs [29], and conservative PINNs [25], etc. In summary, the NRPINN reduces the training cost and improves the prediction accuracy.

There is one limitation in the current work. The NRPINN requires prior informations namely, high-order and zero-order information as the learning tasks of the new Reptile initialization. Therefore, NRPINN is not suitable for solving the problems where the prior information cannot be obtained. In future work, using transfer learning from related works to obtain a initialization may be another way to improve the performance of PINNs.

\section*{Acknowledgement}
This work was supported in part by National Natural Science Foundation of China under Grant No.11725211 and 52005505.

\section*{Compliance with ethical standard Conflict}
\textbf{Conflict of interest} The authors have no conflicts of interest that they are aware of.

\bibliographystyle{spmpsci}

\bibliography{reference}

\end{document}